\documentclass[10pt]{article}

\usepackage[accepted]{tmlr}

\usepackage{amsmath,amsfonts,bm}

\def\eqref#1{equation~\ref{#1}}

\def\1{\bm{1}}

\DeclareMathAlphabet{\mathsfit}{\encodingdefault}{\sfdefault}{m}{sl}
\SetMathAlphabet{\mathsfit}{bold}{\encodingdefault}{\sfdefault}{bx}{n}

\usepackage{microtype}
\usepackage{graphicx}
\usepackage{subcaption}
\usepackage{wrapfig}
\usepackage{booktabs} 

\usepackage{hyperref}
\usepackage{url}

\usepackage{amsmath}
\usepackage{amssymb}
\usepackage{mathtools}
\usepackage{amsthm}
\usepackage{amsfonts}
\usepackage{bbm}
\usepackage{bm}

\usepackage{nicefrac}
\usepackage[table, svgnames, dvipsnames]{xcolor}
\usepackage{tabularx}
\usepackage{longtable}
\usepackage{multirow}
\usepackage{diagbox}
\usepackage{array, makecell}
\usepackage{rotating}
\usepackage{tabu}
\newcommand{\STAB}[1]{\begin{tabular}{@{}c@{}}#1\end{tabular}}
\usepackage{tikz}
\usepackage{pifont}
\usepackage{bbding}

\definecolor{highlight}{HTML}{ffe5ec}

\usepackage[capitalize,noabbrev]{cleveref}

\theoremstyle{plain}

\theoremstyle{definition}

\theoremstyle{remark}

\title{\texttt{OTIS}: Learning High-Quality Time Series Features With Tiny Encoders}

\author{\name Özgün Turgut \email oezguen.turgut@tum.de \\
      \addr Chair for AI in Healthcare and Medicine, Technical University of Munich, Germany \AND
        \name Philip Müller \email philip.j.mueller@tum.de \\
      \addr Chair for AI in Healthcare and Medicine, Technical University of Munich, Germany \AND
        \name Martin J. Menten \email martin.menten@tum.de \\
      \addr Chair for AI in Healthcare and Medicine, Technical University of Munich, Germany \\
            Munich Center for Machine Learning (MCML), Munich, Germany \AND
        \name Daniel Rueckert \email daniel.rueckert@tum.de \\
      \addr Chair for AI in Healthcare and Medicine, Technical University of Munich, Germany \\
            Department of Computing, Imperial College London, UK \\
            Munich Center for Machine Learning (MCML), Munich, Germany}

\def\month{08}  
\def\year{2026} 
\def\openreview{\url{https://openreview.net/forum?id=WW206A1Tru}} 

\begin{document}

\maketitle

\begin{abstract}
We introduce \texttt{OTIS}, an \textbf{o}pen \textbf{ti}me \textbf{s}eries encoder that yields high-quality time series features for downstream deployment on \emph{any} system, including resource-constrained wearables and industrial sensors.\footnote{Code and pre-trained weights are publicly available at \url{https://github.com/oetu/otis}.}
Currently, the development of powerful general-purpose encoders relies on the scaling laws hypothesis, using large encoder sizes to memorise the heterogeneous distributions of multi-domain training data.
However, this reliance on scale creates a barrier to real-world utility, rendering deployment on resource-constrained systems infeasible due to strict memory, energy, and latency constraints.
Surprisingly, we find that tailoring standard masked modelling pre-training to time series properties yields a tiny $7.1\,$M encoder that matches the state-of-the-art performance of $54\times$ larger encoders across $162$ tasks, while requiring $10\times$ less memory, $43\times$ less energy, and $37\times$ lower latency.
To achieve this without the capacity tax, we introduce three novel components:
(1) a \textit{domain-aware tokeniser} to resolve conflicting semantics within multi-domain training data;
(2) a \textit{dual masking strategy} to capture spatiotemporal structures and temporal causality;
and (3) a \textit{structure-aware objective} to decouple feature learning from modelling noise.
Consequently, \texttt{OTIS} produces high-quality time series features that enable state-of-the art performance in discriminative tasks and even extend seamlessly to generative tasks at minimal additional cost.
To democratise access to powerful time series features on any system, we release our code and pre-trained weights.
\end{abstract}

\section{Introduction}
Learning hiqh-quality features from unlabelled data has transformed natural language processing and computer vision (CV).
Driven by self-supervised learning on large-scale data, the field has evolved from specialised encoders to general-purpose encoders capable of solving a variety of tasks across different domains.
In CV, encoders like \texttt{DINOv3} \citep{Simeoni2025} learn visual features (shapes, textures, and depth) that transfer seamlessly across tasks (classification, segmentation, and object detection) and domains (natural, satellite, and medical images).
Recently, this wave has reached time series analysis, yielding two distinct research streams: (i) \textit{forecasting models} such as \texttt{Chronos} \citep{Ansari2024} and \texttt{Time-MoE} \citep{Shi2025} that focus solely on generation, and (ii) \textit{general-purpose encoders} like \texttt{UniTS} \citep{Gao2024} and \texttt{MOMENT} \citep{Goswami2024} that serve discriminative but also simple generative purposes.

However, training a general-purpose time series encoder faces a unique challenge: extreme data \textit{heterogeneity}.
Time series vary heavily across domains in both structure and semantics.
Structural heterogeneity manifests in variate dimensionality, ranging from uni-variate audio to $256$-variate electroencephalography (EEG).
Semantic heterogeneity emerges from distinct variate correlations: EEG variates exhibit physiological connectivities in the brain, differing fundamentally from the thermodynamic laws governing weather variates.
Current approaches, such as \texttt{MOMENT}-$386$M, address this heterogeneity via the \textit{scaling laws hypothesis} \citep{Kaplan2020}, relying on large encoders to memorise data heterogeneity.
While effective, this creates a bottleneck for real-world downstream deployment.
Time series analysis is predominantly required on resource-constrained systems, from wearable health monitors \citep{Shah2025} to industrial edge sensors \citep{Wiesmayr2025}, where deploying large-scale encoders is infeasible due to constraints in memory, energy, and latency.

\begin{figure}[!t]
    \centering
    \includegraphics[width=1.0\linewidth]{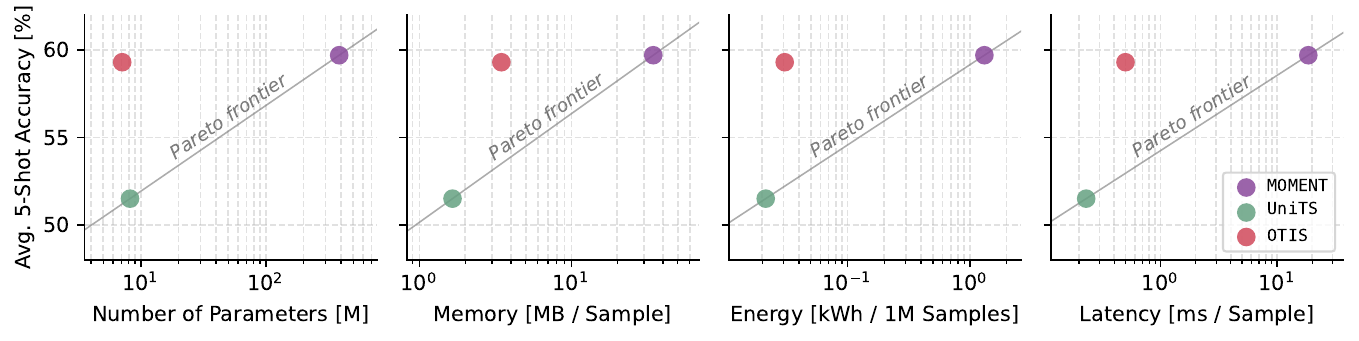}
    \caption{
    \textbf{Extending the Pareto frontier of general-purpose encoders.}
    Across $149$ datasets, \texttt{OTIS} matches the state of the art while requiring $54\times$ fewer parameters, $10\times$ less memory, $43\times$ less energy, and $37\times$ lower latency. 
    This marks a large step towards democratising access to high-quality time series features.
    }
    \label{fig:efficiency}
\end{figure}
In this work, we demonstrate that high-quality time series features can also be learned by tiny encoders via targeted \textit{inductive biases}.
Surprisingly, we find that carefully tailoring standard masked data modelling pre-training to time series properties is sufficient to produce \texttt{OTIS}, a tiny $7.1\,$M \textbf{o}pen \textbf{ti}me \textbf{s}eries encoder that matches state-of-the-art performance across tasks and domains previously reserved for large-scale encoders such as \texttt{MOMENT}-$386$M.
To achieve this with $54\times$ fewer parameters, $10\times$ less memory, $43\times$ less energy, and $37\times$ lower latency (Fig.~\ref{fig:efficiency}), we introduce three novel components:
(1) a \textbf{domain-aware tokeniser} (Sec. \ref{sec:tokeniser}) to resolve conflicting semantics within multi-domain training data via learnable domain-specific embeddings;
(2) a \textbf{dual masking strategy} (Sec. \ref{sec:pre_training}) to model spatiotemporal structures and temporal causality via random and post-fix masking, respectively;
and (3) a \textbf{structure-aware objective} (Sec. \ref{sec:loss}) to decouple feature learning from modelling noise via an additional normalised cross-correlation loss term.
In extensive evaluations on a total of $162$ tasks that go far beyond standard benchmarks, we analyse the deployment of general-purpose encoders in real-world scenarios, including epileptic seizure detection from single-channel EEG and cardiac phenotype regression from $12$-lead electrocardiography (ECG).
These evaluations confirm that \texttt{OTIS} provides high-quality time series features for discriminative tasks, which also seamlessly extend to generative tasks at minimal overhead simply by reusing the lightweight $1.5\,$M decoder from pre-training.

\section{Related Works}
\paragraph{Self-Supervised Learning}
Learning high-quality features from unlabelled data typically relies on carefully designed pretext tasks.
To this end, early approaches leveraged \textit{contrastive learning}, aligning augmented views of a sample in the feature space \citep{Kiyasseh2021, Zhang2022, Woo2022, Wang2022, Radhakrishnan2023, Demirel2023}.
However, identifying effective time series augmentations is challenging and highly domain-dependent.
Consequently, \textit{masked data modelling} (MDM) has emerged as the dominant paradigm.
By reconstructing masked segments from a small visible view, methods like \texttt{PatchTST} \citep{Nie2023}, \texttt{UniTS} \citep{Gao2024}, \texttt{MOMENT} \citep{Goswami2024}, and \texttt{MOIRAI} \citep{Woo2024} learn features without hand-crafted augmentations.
While effective within single domains, standard MDM struggles in multi-domain settings, where data heterogeneity renders learning of high-quality features difficult.

\paragraph{General-Purpose Encoders}
The dominant research stream in time series analysis focuses on \textit{forecasting models} such as \texttt{GPT4TS} \citep{Zhou2023}, \texttt{TTM} \citep{Ekambaram2024}, \texttt{MOIRAI} \citep{Woo2024}, \texttt{Chronos} \citep{Ansari2024}, \texttt{Timer-XL} \citep{Liu2025}, and \texttt{Time-MoE} \citep{Shi2025}.
These works adapt strategies from natural language processing to reliably predict future horizons from past observations, tailoring feature extraction to serve forecasting.
The second stream aims to build \textit{general-purpose encoders} to provide features that serve multiple purposes.
For instance, \texttt{UniTS} \citep{Gao2024} uses specialised task prompts to perform discriminative and generative tasks with a single $8.2\,$M backbone.
However, relying on explicit task prompts risks building fixed mappings for each prompt within the backbone rather than learning generalisable features.
Other approaches such as the $386\,$M \texttt{MOMENT} \citep{Goswami2024} substantially outperform their smaller counterparts simply by exploiting the \textit{scaling laws hypothesis} \citep{Kaplan2020, Hoffmann2022}.
However, relying on scale severely limits deployment on resource-constrained systems like wearables and industrial edge sensors due to strict memory, energy, and latency constraints.

\section{Methodology}
\begin{figure}[t]
    \centering
    \includegraphics[width=0.95\linewidth]{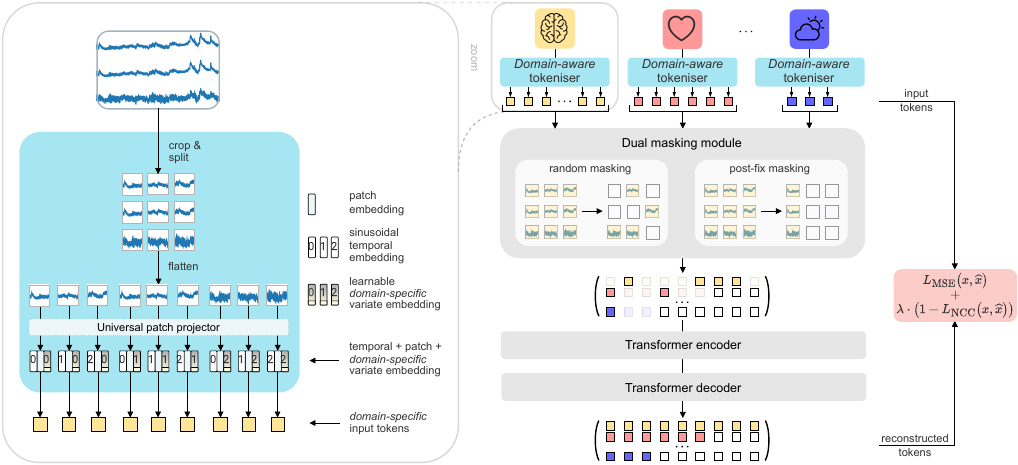}
    \caption{
    \textbf{Effective pre-training on multi-domain data.} Patches are embedded via a universal patch projector.
    Sinusoidal temporal embeddings and learnable domain-specific variate embeddings encode positions.
    Random and post-fix masking is applied to the input, which is reconstructed using a structure-aware loss.
    }
    \label{fig:method}
\end{figure}
In this work, we demonstrate how the extreme heterogeneity of multi-domain training data can be addressed by design rather than scaling, yielding a tiny \textbf{o}pen \textbf{ti}me \textbf{s}eries encoder (\texttt{OTIS}) with state-of-the-art capabilities that is deployable on \textit{any system}.
To build \texttt{OTIS}, we design a simple and robust training recipe based on widely known time series properties.
Particularly, we address multi-domain heterogeneity during tokenisation (Sec.~\ref{sec:tokeniser}) and increase the inductive bias in standard masked data modelling (MDM) pre-training with a simple masking strategy (Sec.~\ref{sec:pre_training}) and regularisation (Sec.~\ref{sec:loss}).
Overall, this simple recipe yields a tiny yet powerful general-purpose encoder that can be easily adapted to \textit{any task} in \textit{any domain} (Sec.~\ref{sec:adaptation}).

\subsection{Domain-Aware Tokenisation}
\label{sec:tokeniser}
Time series data exhibits extreme structural and semantic heterogeneity across domains.
Given a sample $\mathbf{X} \in\mathbb{R}^{V_S \times T}$ from domain $S$, \textit{structural heterogeneity} refers to diverse variate dimensionality ($V_S$), while \textit{semantic heterogeneity} refers to diverse variate correlations (e.g. physiological connectivity versus thermodynamic laws).
This semantic diversity makes pre-training on multi-domain data challenging, forcing the encoder to reconcile conflicting dependencies within a unified representation space.
While \textit{existing methods rely on scaling} encoder and data size to mitigate conflicting learning signals, \textit{we address this by design} to learn powerful time series features with a tiny, highly deployable Transformer \citep{Dosovitskiy2021} encoder.

To this end, we introduce a hierarchical tokenisation strategy that explicitly separates the low-level projection of universal local patterns from the injection of domain-specific semantics.
We first randomly crop or zero-pad $\mathbf{X}$ to a fixed context length $\overline{T}$ and divide it into $T'$ segments of size $P$.
Flattening then yields a sequence of $V_S \cdot T'$ raw time series patches $\mathbf{x}_{v, t} \in \mathbb{R}^{1 \times P}$, with $v \in \{1, \dots, V_S\}$ and $t \in \{1, \dots, T'\}$.

\paragraph{Universal Patch Projection}
We first project the raw patches into a continuous representation space. 
To this end, we use a patch projector consisting of a $1$D convolutional layer with kernel size and stride $P$, followed by layer normalisation and a GELU activation.
This yields \textit{patch embeddings} $\mathbf{e}^\mathcal{P}_{v, t} = e^\mathcal{P}(\mathbf{x}_{v, t}) \in \mathbb{R}^{1 \times D}$, where $D$ denotes the encoder dimension.
As the structure within a single patch arises from local, low-level patterns that occur in any time series regardless of the domain, e.g. valleys, peaks, or saddle points, this projector is kept entirely independent of the source domain, acting as a universal, non-semantic linear projection.

\paragraph{Domain-Aware Spatiotemporal Encoding}
To allow the subsequent time series encoder to build global context and understand the relationships between these isolated patch embeddings, we equip them with spatiotemporal information via decoupled $2$D positional priors.
First, we inject \textit{universal temporal information} using fixed $1$D sinusoidal embeddings $\mathbf{e}^\mathcal{T}_t = e^\mathcal{T}(t) \in \mathbb{R}^{1 \times D}$ to encode the sequential order $t$.
Since the nature of temporal ordering flows consistently across all domains, the temporal embeddings are universally shared.
Second, we inject \textit{domain-specific variate information} using learnable embeddings $\mathbf{e}^\mathcal{V}_{S, v} = e^\mathcal{V}_S(v) \in \mathbb{R}^{1 \times D}$ to encode the variate position $v$.
As the inherent meaning of a single variate and the interaction between multiple variates fundamentally change depending on the source domain $S$, the variate embeddings are kept strictly domain-specific. 
This allows the encoder to learn the unique variate semantics of each distinct domain.

The final input token fed to the encoder is the sum of these three components: $\mathbf{e}_{v, t} = \mathbf{e}^\mathcal{P}_{v, t} + \mathbf{e}^\mathcal{T}_t + \mathbf{e}^\mathcal{V}_{S, v} \in \mathbb{R}^{1 \times D}$, yielding the complete sequence $\mathbf{E} \in \mathbb{R}^{(V_S \cdot T') \times D}$.
This hierarchical tokenisation strategy equips the time series patches with the necessary domain-aware spatiotemporal information, enabling the encoder to focus on dense representation learning without suffering from conflicting learning signals across domains (Fig.~\ref{fig:method}).

\subsection{Dense Representation Learning}
\label{sec:pre_training}
To serve as a general-purpose backbone, a time series encoder requires two distinct capabilities: detecting spatiotemporal structures and modelling temporal causality.
To learn these from unlabelled data, we employ MDM, where a binary mask $\mathbf{m} \in \{0, 1\}^{V_S \cdot T'}$ splits the input sequence $\mathbf{E}$ into visible tokens ($m_{v,t} = 1$) and masked tokens ($m_{v,t} = 0$).
In standard MDM, the randomly masked tokens are reconstructed from the visible view, forcing the encoder to model spatiotemporal structures.
However, random masking allows access to future context, which can impede the learning of temporal causality.
We therefore propose \textit{dual masking}, a strategy that alternates between two distinct distributions of $m_{v,t}$ during pre-training (App.~\ref{sec:app_dual_masking}).

\begin{wrapfigure}[10]{r}{0.5\linewidth}
    \centering
    \vspace{-5pt}
    \includegraphics[width=\linewidth]{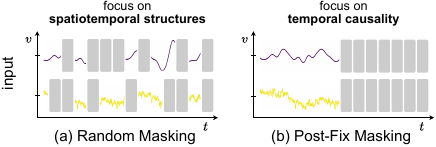}
    \caption{
    \textbf{Learning spatiotemporal structures and temporal causality} via (a) random masking and (b) post-fix masking.
    }
    \label{fig:masking_strategy}
\end{wrapfigure}

\paragraph{Structural Learning}
In $75\,\%$ of training steps, we apply \textit{random masking} where $m_{v, t} \sim \text{Bernoulli}(1 - \rho)$, with $\rho \in [0, 1]$ denoting the masking
ratio.
This forces the encoder to learn the spatiotemporal structures necessary for accurate reconstruction (Fig.~\ref{fig:masking_strategy}a).

\paragraph{Causal Learning}
In the remaining $25\,\%$ of steps, we apply \textit{post-fix masking} where $m_{v, t} = \mathbbm{1}_{[t \leq T'/2]}$.
Hiding future context transforms the reconstruction task into a forecasting task, explicitly forcing the encoder to model temporal causality (Fig.~\ref{fig:masking_strategy}b).

Once a mask $\mathbf{m}$ is drawn, it is applied to the input $\mathbf{E}$, yielding a reduced sequence $\mathbf{E}[\mathbf{m}] \in \mathbb{R}^{N_\text{v} \times D}$, where $N_\text{v}$ denotes the number of visible tokens.
The encoder $f(\cdot)$ processes this sequence to produce token-level time series features:
\begin{equation}
\mathbf{H} = f(\mathbf{E}[\mathbf{m}]) \in \mathbb{R}^{N_\text{v} \times D}.
\end{equation}
To reconstruct the input, a full sequence $\mathbf{H}' \in \mathbb{R}^{(V_S \cdot T') \times D}$ is created by combining the token-level features with learnable mask embeddings $\mathbf{e}^\mathcal{M} \in \mathbb{R}^{1 \times D}$ that are inserted at the masked positions.
Crucially, we re-inject the positional embeddings to provide exact spatiotemporal context to the decoder:
\begin{equation}
\mathbf{h}'_{v, t} = \begin{cases}
    \mathbf{h}_{v, t} + \mathbf{e}^\mathcal{T}_t + \mathbf{e}^\mathcal{V}_{S, v} & \text{if } m_{v, t} = 1 \\
    \mathbf{e}^\mathcal{M} + \mathbf{e}^\mathcal{T}_t + \mathbf{e}^\mathcal{V}_{S, v} & \text{if } m_{v, t} = 0
\end{cases} \quad \in \mathbb{R}^{1 \times D}.
\end{equation}
The decoder $g(\cdot)$ then reconstructs the input patches:
\begin{equation}
\mathbf{\widehat{X}} = g(\mathbf{H}') \in \mathbb{R}^{V_S \times (T' \cdot P)},
\end{equation}
where $(T' \cdot P) = \overline{T}$ is the total context length in time points.

\subsection{Structure-Aware Optimisation}
\label{sec:loss}
\begin{wrapfigure}{r}
{0.5\linewidth}
    \centering
    \vspace{0pt}
    \includegraphics[width=\linewidth]{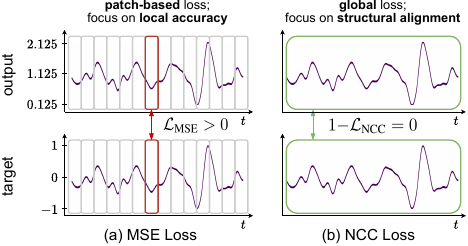}
    \caption{
    \textbf{Optimising local accuracy and global structure} using an (a) MSE loss and (b) NCC loss.
    }
    \label{fig:loss_function}
\end{wrapfigure}

Standard MDM typically optimises reconstruction using a mean squared error (MSE) loss, which measures the point-wise Euclidean distance between the input and the reconstruction:
\begin{equation}
\label{eq:mse}
\mathcal{L}_\text{MSE} = \frac{1}{V_S \cdot T'}\sum_{v=1}^{V_S} \sum_{t=1}^{T'} \| \mathbf{x}_{v, t} - \widehat{\mathbf{x}}_{v, t} \|_2^2.
\end{equation}
While effective for enforcing \textit{local accuracy}, relying solely on MSE is suboptimal for learning robust time series features, as it treats time points independently and ignores the global waveform structure.
Consequently, reconstructions that capture the correct shape but suffer from amplitude scaling or offset shifts incur disproportionately high penalties (Fig.~\ref{fig:loss_function}a).
Furthermore, when faced with uncertainty in masked regions, MSE biases predictions toward the local mean \citep{Mathieu2016, Isola2017, Rahaman2019}, resulting in overly smooth reconstructions that lack structural fidelity.

To mitigate this, we introduce a \textit{normalised cross-correlation} (NCC) loss.
By measuring the linear dependence between the input and the reconstruction, it rewards relative alignment in phase and shape rather than strict absolute accuracy (Fig.~\ref{fig:loss_function}b):
\begin{equation}
\label{eq:ncc_loss}
\mathcal{L}_{\text{NCC}} = \frac{1}{V_S \cdot \overline{T}} \sum^{V_S}_{v=1} \sum^{\overline{T}}_{i=1} \frac{(x_{v,i} - \mu_{\mathbf{x}_v}) (\widehat{x}_{v,i} - \mu_{\widehat{\mathbf{x}}_v})}{\sigma_{\mathbf{x}_v} \sigma_{\widehat{\mathbf{x}}_v}},
\end{equation}
where $x_{v,i}$ denotes the value at the $i$-th time point of variate $v$, while $\mu$ and $\sigma$ represent the variate-level mean and standard deviation, respectively.
Maximising NCC preserves the \textit{global structure} and prevents potential mean-collapse.
The final objective balances local accuracy and structural alignment:
\begin{equation}
\label{eq:total_loss}
\mathcal{L} = \mathcal{L}_{\text{MSE}} + \lambda \cdot (1 - \mathcal{L}_{\text{NCC}}),
\end{equation}
where $\lambda$ is a weighting hyperparameter.
This composite objective allows the encoder to learn meaningful time series features without wasting modelling capacity on noise or scaling artifacts.

\subsection{Versatile Downstream Deployment}
\label{sec:adaptation}

The primary use of \texttt{OTIS} is the extraction of high-quality time series features for \textit{discriminative} tasks, including classification and regression. 
To this end, the pre-trained encoder $f(\cdot)$ processes an unmasked input time series $\mathbf{E}$ to extract \textit{features} $\mathbf{H} = f(\mathbf{E}) \in \mathbb{R}^{(V_S \cdot T') \times D}$. 
These token-level features are subsequently aggregated via global average pooling to yield a dense \textit{representation} $\mathbf{h}^* \in \mathbb{R}^{1 \times D}$. 
Due to its tiny size of $7.1\,$M parameters, \texttt{OTIS} enables both zero-shot inference on a CPU and fine-tuning on a single commodity GPU. 
Crucially, utilising randomly initialised variate embeddings $\mathbf{e}^\mathcal{V}_{S, v} = e^\mathcal{V}_S(v) \in \mathbb{R}^{1 \times D}$ for tasks in previously unseen domains is sufficient for competitive performance (App. \ref{sec:app_performance} and \ref{sec:app_kshot}).

Beyond discriminative tasks, \texttt{OTIS} features $\mathbf{H}$ can be seamlessly repurposed for simple \textit{generative} tasks, such as short-term forecasting, with minimal computational overhead. 
To this end, the lightweight $1.5\,$M decoder $g(\cdot)$ retained from pre-training is appended to the encoder $f(\cdot)$.
For optimal generative performance, both modules can be fine-tuned end-to-end utilising post-fix masking and the composite loss defined in Eq.~\ref{eq:total_loss}.

\section{Experiments}
\label{sec:experiments}
\paragraph{Architecture}
\texttt{OTIS} is a tiny $7.1\,$M Transformer \citep{Vaswani2017} encoder $f(\cdot)$ with $12$ layers, a dimension of $192$, an MLP size of $768$, and $3$ heads.
For the masked data modelling objective, a lightweight $1.5\,$M Transformer decoder $g(\cdot)$ with $4$ layers, a dimension of $160$, an MLP size of $640$, and $5$ heads is employed.
Patch size and stride are set to $P=24$, splitting a time series into $T'=\overline{T}/P$ non-overlapping patches along the temporal axis.
During pre-training, a context length of $\overline{T}=1008$ time points is used, resulting in $T'=42$ temporal embeddings.
Crucially, for inference, the temporal embeddings are linearly interpolated to match any sequence length, providing the downstream flexibility to process time series of arbitrary duration.

\paragraph{Data \& Training}
We curate a diverse pre-training corpus ($640\,$k samples, $11\,$B time points; App.~\ref{sec:app_ts_corpus}) spanning healthcare, engineering, natural sciences, and finance.
This diversity ensures \texttt{OTIS} is exposed to heterogeneous variate semantics and temporal dynamics.
Particularly, to align with the growing interest in deploying general-purpose backbones for healthcare applications \citep{Saab2024, Zhang2024NMed, Turgut2025EEG}, our corpus includes large datasets of the most widely used physiological signals, $12$-lead ECG and $10$-$20$ system EEG.
We pre-train \texttt{OTIS} for $200$ epochs on $4$ NVIDIA A$100$ GPUs using AdamW \citep{Loshchilov2017} with batch size of $5120$, masking ratio $\rho=0.75$, NCC $\lambda=0.1$ (App.~\ref{sec:app_ncc}), and cosine annealing with learning rate of $3e$-$5$ and $10\,\%$ warm-up.
Fine-tuning and inference run on one NVIDIA A$6000$ GPU.
Details on computational costs and hyperparameters are provided in Appendices \ref{sec:app_comp_cost} and \ref{sec:app_exp_details}.

\paragraph{Benchmarks}
We evaluate \texttt{OTIS} on a total of $162$ downstream tasks (App.~\ref{sec:app_benchmark_details}).
To test its \textit{discriminative capabilities}, we evaluate in two key areas.
First, standard classification tasks including fault detection in rolling bearings (FD-B \citet{Lessmeier2016}), hand-gestures from accelerometers (Gesture \citet{Liu2009}), and muscular diseases from electromyography (EMG \citet{Goldberger2000}).
We follow standard procedures from the original works and report accuracy (ACC $\uparrow$).
We further benchmark on $23$ UEA \citep{Bagnall2018} and $121$ UCR \citep{Dau2019} archive datasets, following standard protocols and reporting balanced accuracy (bACC $\uparrow$).
Second, to specifically validate potential deployment on wearable health monitors, we extend the evaluation to real-world medical tasks.
These include the detection of epileptic seizures (Epilepsy \citet{Andrzejak2001}), sleep stages (SleepEDF \citet{Goldberger2000}), and event types (TUEV \citet{Harati2015}) from EEG.
We follow standard procedures from the original works and report balanced accuracy (bACC $\uparrow$).
Further, we predict cardiac phenotypes such as the left ventricular (LV) end-diastolic volume (EDV), end-systolic volume (ESV), stroke volume (SV), ejection fraction (EF), and mass (M) from ECG (UK BioBank \citep{Sudlow2015}).
We follow standard protocols from \citet{Bai2024} and \citet{Turgut2025} and report R-squared ($R^2$ $\uparrow$).
To explore the \textit{generative capabilities} of \texttt{OTIS}, we reuse the lightweight $1.5\,$M decoder retained from pre-training and perform short-term forecasting (look-back window of $96$ time points and prediction horizon of $336$ time points).
These include standard benchmarks on transformer temperature (ETT \citet{Zhou2021}), meteorology (Weather \citet{Wetter2024}), household electricity consumption (Electricity \citet{UCI2024}), influenza-like illness occurrence (ILI \citep{ILI2024}), and road occupancy (Traffic \citet{PeMS2024}).
We follow established procedures from \citet{Zhou2021} and report normalised mean squared error (MSE $\downarrow$).
Mean and standard deviation are reported across five seeds set during fine-tuning.
 
\paragraph{Baselines}
We compare \texttt{OTIS} against a comprehensive suite of $31$ baselines (App.~\ref{sec:app_baseline_details}).
Primarily, we select state-of-the-art \textit{general-purpose encoders} pre-trained on large heterogeneous corpora, specifically the $8.2\,$M \texttt{UniTS} \citep{Gao2024} and $386\,$M \texttt{MOMENT} \citep{Goswami2024}.
We also include \textit{simple encoders} for general time series analysis like \texttt{TS2Vec} \citep{Yue2022}, \texttt{TimesNet} \citep{Wu2022}, \texttt{TF-C} \citep{Zhang2022}, \texttt{Ti-MAE} \citep{Li2023}, \texttt{SimMTM} \citep{Dong2023}, \texttt{PatchTST} \citep{Nie2023}, \texttt{CroSSL} \citep{Deldari2024}, \texttt{TSLANet} \citep{Eldele2024}, and \texttt{InvConvNet} \citep{Germain2025}.
For the medical benchmarks, we compare against \textit{specialised EEG encoders}, such as \texttt{Handcrafted} \citep{Engemann2022}, \texttt{ShallowNet} \citep{Schirrmeister2017}, \texttt{DeepNet} \citep{Schirrmeister2017}, \texttt{EEGNet} \citep{Lawhern2018}, \texttt{CBraMod} \citep{Wang2025CBra}, and \texttt{CSBrain} \citep{Zhou2025}.
Further, we evaluate against \textit{specialised ECG encoders} trained on UK BioBank \citep{Sudlow2015}, the largest biomedical databank, including a supervised Transformer, \texttt{ECG-Transformer}, a Transformer pre-trained with masked data modelling \citep{He2022}, \texttt{ECG-MAE}, and Transformers trained multi-modally with paired imaging data, \texttt{CM-AE} \citep{Radhakrishnan2023} and \texttt{MMCL} \citep{Turgut2025}.
For the exploratory forecasting experiments, we include \textit{simple models} for forecasting such as \texttt{N-BEATS} \citep{Oreshkin2019} and \texttt{DLinear} \citep{Zeng2023}, as well as \textit{specialised forecasting models} ranging from the $5\,$M \texttt{TTM} \citep{Ekambaram2024} to the $2.4\,$B \texttt{Time-MoE} \citep{Shi2025}, alongside \texttt{GPT4TS} \citep{Zhou2023}, \texttt{MOIRAI} \citep{Woo2024}, \texttt{Chronos} \citep{Ansari2024}, \texttt{ROSE} \citep{Wang2025rose}, \texttt{Timer-XL} \citep{Liu2025}, and \texttt{DTAF} \citep{Lu2026}.

\section{Results}
\label{sec:results}
\subsection{\texttt{OTIS} Is Competitive with State of the Art \textit{and} Highly Deployable}
We assess whether \texttt{OTIS} (i) is competitive with state-of-the-art specialised and general-purpose encoders, (ii)~remains deployable under strict hardware constraints including memory, energy, and latency, and (iii)~produces features suitable for simple generative tasks.

\begin{table}[!t]
\caption{
\textbf{Classification.}
\textbf{Best} and \underline{second best} scores (ACC $\uparrow$) are highlighted.
\texttt{OTIS} demonstrates strong performance, outperforming specialised encoders as well as large general-purpose encoders.
}
\label{tab:discriminative}
\begin{subtable}[t]{0.49\linewidth}
    \caption{Standard benchmarks.}
    \label{tab:classic_cls}
    \centering
    \footnotesize
    \setlength{\tabcolsep}{0.75em}
    \begin{tabular}{lrcccc}
        \toprule
        \textbf{Encoder} & \textbf{\makecell{Machine\\fault}} & \textbf{\makecell{Hand\\gesture}} & \textbf{\makecell{Muscular\\disease}} \\
        \midrule
        \multicolumn{4}{l}{\footnotesize\textit{\textcolor{lightgray}{Simple encoders}}} \\
        \texttt{TS2Vec} & 0.479 \scalebox{0.65}{$\pm$ 0.011} & 0.692 \scalebox{0.65}{$\pm$ 0.033} & 0.785 \scalebox{0.65}{$\pm$ 0.032} \\
        \texttt{TimesNet} & 0.619 \scalebox{0.65}{$\pm$ 0.020} & 0.598 \scalebox{0.65}{$\pm$ 0.026} & 0.912 \scalebox{0.65}{$\pm$ 0.021} \\
        \texttt{TF-C} & 0.694 \scalebox{0.65}{$\pm$ 0.023} & 0.764 \scalebox{0.65}{$\pm$ 0.019} & 0.817 \scalebox{0.65}{$\pm$ 0.029} \\
        \texttt{Ti-MAE} & 0.609 \scalebox{0.65}{$\pm$ 0.031} & 0.719 \scalebox{0.65}{$\pm$ 0.023} & 0.700 \scalebox{0.65}{$\pm$ 0.017} \\
        \texttt{SimMTM} & 0.685 \scalebox{0.65}{$\pm$ 0.042} & 0.778 \scalebox{0.65}{$\pm$ 0.074} & 0.951 \scalebox{0.65}{$\pm$ 0.000} \\
        \texttt{PatchTST} & 0.730 \scalebox{0.65}{$\pm$ 0.052} & 0.617 \scalebox{0.65}{$\pm$ 0.000} & 0.932 \scalebox{0.65}{$\pm$ 0.032} \\
        \texttt{TSLANet} & 0.756 \scalebox{0.65}{$\pm$ 0.020} & \textbf{0.787} \scalebox{0.65}{$\pm$ 0.014} & 0.951 \scalebox{0.65}{$\pm$ 0.000} \\
        \texttt{CroSSL} & 0.737 \scalebox{0.65}{$\pm$ 0.026} & 0.610 \scalebox{0.65}{$\pm$ 0.014} & 0.951 \scalebox{0.65}{$\pm$ 0.000} \\
        \texttt{InvConvNet} & 0.681 \scalebox{0.65}{$\pm$ 0.009} & 0.458 \scalebox{0.65}{$\pm$ 0.031} & 0.961 \scalebox{0.65}{$\pm$ 0.033} \\
        \midrule
        \multicolumn{4}{l}{\footnotesize\textit{\textcolor{lightgray}{General-purpose encoders}}} \\
        \texttt{MOMENT-386M} & \underline{0.928} \scalebox{0.65}{$\pm$ 0.024} & 0.618 \scalebox{0.65}{$\pm$ 0.017} & \underline{0.976} \scalebox{0.65}{$\pm$ 0.000} \\
        \texttt{UniTS-8.2M} & 0.749 \scalebox{0.65}{$\pm$ 0.052} & 0.615 \scalebox{0.65}{$\pm$ 0.009} & 0.971 \scalebox{0.65}{$\pm$ 0.011} \\
        \rowcolor{highlight}
        \texttt{OTIS-7.1M} & \textbf{0.981} \scalebox{0.65}{$\pm$ 0.007} & \underline{0.780} \scalebox{0.65}{$\pm$ 0.020} & \textbf{0.976} \scalebox{0.65}{$\pm$ 0.000} \\
        \bottomrule
    \end{tabular}
\end{subtable}
\hfill
\begin{subtable}[t]{0.49\linewidth}
    \centering
    \caption{EEG benchmarks.}
    \label{tab:eeg_cls}
    \footnotesize
    \setlength{\tabcolsep}{0.6em}
    \begin{tabular}{lrccccc}
        \toprule
        \textbf{Encoder} & \textbf{\makecell{Epileptic\\seizure}} & \textbf{\makecell{Sleep\\stage}} & \textbf{\makecell{Event\\type}} \\
        \midrule
        \multicolumn{4}{l}{\footnotesize\textit{\textcolor{lightgray}{Simple encoders}}} \\
        \texttt{PatchTST} & 0.861 \scalebox{0.65}{$\pm$ 0.021} & 0.699 \scalebox{0.65}{$\pm$ 0.033} & 0.531 \scalebox{0.65}{$\pm$ 0.212} \\
        \texttt{TSLANet} & 0.899 \scalebox{0.65}{$\pm$ 0.010} & 0.726 \scalebox{0.65}{$\pm$ 0.011} & 0.460 \scalebox{0.65}{$\pm$ 0.031} \\
        \texttt{CroSSL} & 0.862 \scalebox{0.65}{$\pm$ 0.022} & 0.701 \scalebox{0.65}{$\pm$ 0.005} & 0.380 \scalebox{0.65}{$\pm$ 0.036} \\
        \texttt{InvConvNet} & 0.854 \scalebox{0.65}{$\pm$ 0.040} & 0.675 \scalebox{0.65}{$\pm$ 0.016} & 0.449 \scalebox{0.65}{$\pm$ 0.024} \\
        \multicolumn{4}{l}{\footnotesize\textit{\textcolor{lightgray}{Specialised EEG encoders}}} \\
        \texttt{Handcrafted} & \textbf{0.930} \scalebox{0.65}{$\pm$ 0.000} & 0.655 \scalebox{0.65}{$\pm$ 0.000} & 0.311 \scalebox{0.65}{$\pm$ 0.000} \\
        \texttt{ShallowNet} & 0.901 \scalebox{0.65}{$\pm$ 0.021} & 0.687 \scalebox{0.65}{$\pm$ 0.015} & 0.493 \scalebox{0.65}{$\pm$ 0.026} \\
        \texttt{DeepNet} & 0.899 \scalebox{0.65}{$\pm$ 0.011} & 0.740 \scalebox{0.65}{$\pm$ 0.007} & 0.561 \scalebox{0.65}{$\pm$ 0.017} \\
        \texttt{EEGNet} & 0.872 \scalebox{0.65}{$\pm$ 0.019} & 0.708 \scalebox{0.65}{$\pm$ 0.008} & 0.474 \scalebox{0.65}{$\pm$ 0.015} \\
        \texttt{CBraMod} & 0.858 \scalebox{0.65}{$\pm$ 0.011} & 0.735 \scalebox{0.65}{$\pm$ 0.004} & 0.472 \scalebox{0.65}{$\pm$ 0.019} \\
        \texttt{CSBrain} & 0.855 \scalebox{0.65}{$\pm$ 0.029} & 0.720 \scalebox{0.65}{$\pm$ 0.019} & 0.522 \scalebox{0.65}{$\pm$ 0.030} \\
        \midrule
        \multicolumn{4}{l}{\footnotesize\textit{\textcolor{lightgray}{General-purpose encoders}}} \\
        \texttt{MOMENT-386M} & 0.918 \scalebox{0.65}{$\pm$ 0.007} & \textbf{0.756} \scalebox{0.65}{$\pm$ 0.002} & 0.517 \scalebox{0.65}{$\pm$ 0.044} \\
        \texttt{UniTS-8.2M} & 0.878 \scalebox{0.65}{$\pm$ 0.012} & 0.710 \scalebox{0.65}{$\pm$ 0.011} & \underline{0.546} \scalebox{0.65}{$\pm$ 0.021} \\
        \rowcolor{highlight}
        \texttt{OTIS-7.1M} & \underline{0.918} \scalebox{0.65}{$\pm$ 0.014} & \underline{0.749} \scalebox{0.65}{$\pm$ 0.006} & \textbf{0.614} \scalebox{0.65}{$\pm$ 0.013} \\
        \bottomrule
    \end{tabular}
\end{subtable}
\end{table}

\begin{table}[t]
\caption{
\textbf{Regression.}
\textbf{Best} and \underline{second best} scores ($R^2$ $\uparrow$) are highlighted.
\texttt{OTIS} features are robust for critical healthcare applications such as cardiac phenotype regression from standard $12$-lead electrocardiography.
}
\label{tab:regression}
\centering
\footnotesize
\setlength{\tabcolsep}{1.0em}
\begin{tabular}{lcccccc}
    \toprule
    \textbf{Encoder} & \textbf{LVEDV} & \textbf{LVESV} & \textbf{LVSV} & \textbf{LVEF} & \textbf{LVM} \\
    \midrule
    \multicolumn{4}{l}{\footnotesize\textit{\textcolor{lightgray}{Simple encoders}}} \\
    \texttt{PatchTST} & 0.321 \scalebox{0.65}{$\pm$ 0.006} & 0.279 \scalebox{0.65}{$\pm$ 0.015} & 0.234 \scalebox{0.65}{$\pm$ 0.006} & 0.063 \scalebox{0.65}{$\pm$ 0.012} & 0.369 \scalebox{0.65}{$\pm$ 0.042} \\
    \texttt{ECG-Transformer} & 0.394 \scalebox{0.65}{$\pm$ 0.008} & 0.386 \scalebox{0.65}{$\pm$ 0.015} & 0.294 \scalebox{0.65}{$\pm$ 0.011} & 0.171 \scalebox{0.65}{$\pm$ 0.007} & 0.459 \scalebox{0.65}{$\pm$ 0.021} \\
    \texttt{TSLANet} & 0.447 \scalebox{0.65}{$\pm$ 0.008} & 0.421 \scalebox{0.65}{$\pm$ 0.007} & 0.327 \scalebox{0.65}{$\pm$ 0.008} & 0.173 \scalebox{0.65}{$\pm$ 0.008} & 0.513 \scalebox{0.65}{$\pm$ 0.012} \\
    \texttt{CroSSL} & 0.451 \scalebox{0.65}{$\pm$ 0.004} & 0.418 \scalebox{0.65}{$\pm$ 0.003} & 0.315 \scalebox{0.65}{$\pm$ 0.003} & 0.123 \scalebox{0.65}{$\pm$ 0.003} & 0.506 \scalebox{0.65}{$\pm$ 0.002} \\
    \texttt{InvConvnet} & 0.369 \scalebox{0.65}{$\pm$ 0.001} & 0.325 \scalebox{0.65}{$\pm$ 0.001} & 0.267 \scalebox{0.65}{$\pm$ 0.001} & 0.082 \scalebox{0.65}{$\pm$ 0.001} & 0.430 \scalebox{0.65}{$\pm$ 0.002} \\
    \multicolumn{4}{l}{\footnotesize\textit{\textcolor{lightgray}{Specialised ECG encoders}}} \\
    \texttt{ECG-MAE} & 0.478 \scalebox{0.65}{$\pm$ 0.013} & 0.475 \scalebox{0.65}{$\pm$ 0.010} & 0.349 \scalebox{0.65}{$\pm$ 0.002} & 0.233 \scalebox{0.65}{$\pm$ 0.016} & 0.565 \scalebox{0.65}{$\pm$ 0.008} \\
    \texttt{CM-AE}* & 0.451 \scalebox{0.65}{$\pm$ 0.006} & 0.380 \scalebox{0.65}{$\pm$ 0.019} & 0.316 \scalebox{0.65}{$\pm$ 0.011} & 0.103 \scalebox{0.65}{$\pm$ 0.012} & 0.536 \scalebox{0.65}{$\pm$ 0.016} \\
    \texttt{MMCL}* & \underline{0.498} \scalebox{0.65}{$\pm$ 0.012} & \underline{0.497} \scalebox{0.65}{$\pm$ 0.008} & \underline{0.360} \scalebox{0.65}{$\pm$ 0.013} & \underline{0.245} \scalebox{0.65}{$\pm$ 0.004} & \underline{0.597} \scalebox{0.65}{$\pm$ 0.018} \\
    \midrule
    \multicolumn{4}{l}{\footnotesize\textit{\textcolor{lightgray}{General-purpose encoders}}} \\
    \texttt{MOMENT-386M} & 0.449 \scalebox{0.65}{$\pm$ 0.012} & 0.431 \scalebox{0.65}{$\pm$ 0.022} & 0.344 \scalebox{0.65}{$\pm$ 0.014} & 0.181 \scalebox{0.65}{$\pm$ 0.032} & 0.532 \scalebox{0.65}{$\pm$ 0.012} \\
    \texttt{UniTS-8.2M} & 0.479 \scalebox{0.65}{$\pm$ 0.013} & 0.458 \scalebox{0.65}{$\pm$ 0.011} & 0.355 \scalebox{0.65}{$\pm$ 0.021} & 0.197 \scalebox{0.65}{$\pm$ 0.024} & 0.555 \scalebox{0.65}{$\pm$ 0.011} \\
    \rowcolor{highlight}
    \texttt{OTIS-7.1M} & \textbf{0.506} \scalebox{0.65}{$\pm$ 0.002} & \textbf{0.498} \scalebox{0.65}{$\pm$ 0.001} & \textbf{0.386} \scalebox{0.65}{$\pm$ 0.004} & \textbf{0.253} \scalebox{0.65}{$\pm$ 0.001} & \textbf{0.604} \scalebox{0.65}{$\pm$ 0.002} \\
    \bottomrule
    \multicolumn{6}{l}{\footnotesize{*Multi-modally pre-trained on paired time series and imaging data.}} \\
\end{tabular}
\end{table}
\paragraph{Discriminative Capabilities}
On standard benchmarks (Tab.~\ref{tab:classic_cls}), \texttt{OTIS} shows strong discriminative capabilities, consistently outperforming the general-purpose encoders \texttt{UniTS} and \texttt{MOMENT}.
Notably, it achieves state-of-the-art on FD-B despite the dataset containing long time series ($T=5120$).
As \texttt{OTIS} was pre-trained with $\overline{T}=1008$, this confirms that our linear interpolation of temporal embeddings generalises to unseen resolutions and sequence lengths.
The encoder adapts seamlessly to novel domains such as inertial measurement units in wearables (Gesture) and single-channel EMG, suggesting that it does not overfit to the distributions of the pre-training data but rather learns a meaningful representation space.
This quality extends to medical benchmarks, particularly electroencephalography (EEG), where \texttt{OTIS} outperforms specialised EEG encoders (Tab.~\ref{tab:eeg_cls}).
Further, on electrocardiography (ECG) regression (Tab.~\ref{tab:regression}), it consistently surpasses specialised ECG encoders such as \texttt{MMCL} and \texttt{CM-AE} which use multi-modal supervision.
These results confirm that \texttt{OTIS} captures predictive \textit{spatiotemporal structures} robust enough for critical healthcare applications.

\paragraph{Frozen-Encoder Capabilities}
To evaluate the quality of the learned time series features, we  \textit{freeze} \texttt{OTIS} directly after pre-training and probe it on $144$ unseen datasets, including $23$ from the UEA multi-variate archive \citep{Bagnall2018} and $121$ from the UCR uni-variate archive \citep{Dau2019}.
In particular, we utilise three distinct protocols: prototypical $5$-shot classification, $k$-nearest-neighbour classification, and linear probing (see App.~\ref{sec:app_performance} for full details).
Beyond performance, we evaluate its deployability on a commodity GPU, including memory, energy, and latency footprints as well as throughput (see App.~\ref{sec:app_deployment} for full details).
On multi-variate time series (Tab.~\ref{tab:kshot_avg} UEA), we find \texttt{OTIS} to achieve the best performance compared against the current state of the art ($55.7\,\%$ $5$-shot / $60.8\,\%$ k-NN / $68.1\,\%$ linear versus $51.5$/$55.4$/$64.7\,\%$ \texttt{MOMENT}, $46.3$/$50.4$/$56.9\,\%$ \texttt{UniTS}).
Most notably, our tiny encoder wins on $16$ of $23$ UEA datasets, while consuming $6\!\times$ less memory than the currently best \texttt{MOMENT} ($11.5$ vs $66.3$\,MB/sample), $45\!\times$ less energy ($0.1$ vs $4.5$\,kWh/1M\,samples), and operating at $46\!\times$ lower latency ($1.4$ vs $64.3$\,ms/sample).
On uni-variate time series (Tab.~\ref{tab:kshot_avg} UCR), it remains competitive ($59.8\,\%$~$5$-shot / $60.5\,\%$ k-NN~/ $73.7\,\%$ linear versus $60.9$/$61.4$/$75.8\,\%$ \texttt{MOMENT}, $52.2$/$56.4$/$67.9\,\%$ \texttt{UniTS}) at $25\!\times$ lower memory than \texttt{MOMENT} ($1.1$ vs $27.3$\,MB/sample), $31\!\times$ less energy ($0.014$ vs $0.437$\,kWh/1M\,samples), and $20\!\times$ lower latency ($0.3$ vs $6.3$\,ms/sample).
Consequently, we find \texttt{OTIS} to yield substantial memory ($-90\,\%$), energy ($-98\,\%$), and latency ($-97\,\%$) gains against the current state of the art (Tab.~\ref{tab:kshot_overall}), while remaining highly competitive across a diverse set of tasks and domains.

\begin{table}[!t]
\caption{
\textbf{Frozen-encoder classification on the UEA and UCR archives.}
\textbf{Best} and \underline{second best} $5$-shot, $k$-nearest-neighbour, and linear probing results (bACC in \% $\uparrow$) are highlighted.
Wins out of total datasets are reported for 5-shot classification.
\texttt{OTIS} attains the highest scores on the UEA archive and remains competitive with the $54\!\times$ larger \texttt{MOMENT} on the UCR archive, at a small fraction of its memory, energy, and latency cost.
}
\label{tab:kshot_avg}
\centering
\footnotesize
\setlength{\tabcolsep}{0.25em}
\begin{tabular}{llcccccccc}
    \toprule
    \textbf{Archive} & \textbf{Encoder} & \textbf{$5$-Shot} & \textbf{k-NN} & \textbf{Linear} & \textbf{Wins} & \textbf{\makecell{Memory\\{\tiny [MB/Sample]} $\downarrow$}} & \textbf{\makecell{Energy\\{\tiny [kWh/1M\,Samples]} $\downarrow$}} & \textbf{\makecell{Latency\\{\tiny [ms/Sample]} $\downarrow$}} & \textbf{\makecell{Throughput\\{\tiny [Samples/s]} $\uparrow$}} \\
    \midrule
    \multirow{3}{*}{UEA} & \texttt{MOMENT-386M} & \underline{$51.5$} \scalebox{0.65}{$\pm$ 2.8} & \underline{$55.4$} \scalebox{0.65}{$\pm$ 4.9} & \underline{$64.7$} \scalebox{0.65}{$\pm$ 5.7} & \underline{$5$} & $66.3$ & $4.5$ & $64.3$ & $168.2$ \\
    & \texttt{UniTS-8.2M} & $46.3$ \scalebox{0.65}{$\pm$ 3.0} & $50.4$ \scalebox{0.65}{$\pm$ 4.1} & $56.9$ \scalebox{0.65}{$\pm$ 5.3} & $2$ & $4.9$ & $0.1$ & $0.8$ & $5149.9$ \\
    \rowcolor{highlight}
    \cellcolor{white} & \texttt{OTIS-7.1M} & $\mathbf{55.7}$ \scalebox{0.65}{$\pm$ 2.6} & $\mathbf{60.8}$ \scalebox{0.65}{$\pm$ 4.8} & $\mathbf{68.1}$ \scalebox{0.65}{$\pm$ 5.4} & $\mathbf{16}$ & $11.5$ & $0.1$ & $1.4$ & $2549.4$ \\
    \midrule
    \multirow{3}{*}{UCR} & \texttt{MOMENT-386M} & $\mathbf{60.9}$ \scalebox{0.65}{$\pm$ 3.3} & $\mathbf{61.4}$ \scalebox{0.65}{$\pm$ 2.6} & $\mathbf{75.8}$ \scalebox{0.65}{$\pm$ 2.4} & $\mathbf{63}$ & $27.3$ & $0.437$ & $6.3$ & $492.2$ \\
    & \texttt{UniTS-8.2M} & $52.2$ \scalebox{0.65}{$\pm$ 3.5} & $56.4$ \scalebox{0.65}{$\pm$ 3.1} & $67.9$ \scalebox{0.65}{$\pm$ 2.5} & $11$ & $0.9$ & $0.005$ & $0.1$ & $12238.2$ \\
    \rowcolor{highlight}
    \cellcolor{white} & \texttt{OTIS-7.1M} & \underline{$59.8$} \scalebox{0.65}{$\pm$ 3.4} & \underline{$60.5$} \scalebox{0.65}{$\pm$ 2.3} & \underline{$73.7$} \scalebox{0.65}{$\pm$ 2.4} & \underline{$47$} & $1.1$ & $0.014$ & $0.3$ & $3850.6$ \\
    \bottomrule
\end{tabular}
\end{table}

\begin{table}[t]
\centering
\caption{
\textbf{Forecasting.}
\textbf{Best} and \underline{second best} scores (MSE $\downarrow$) are highlighted.
The utility of \texttt{OTIS} time series features seamlessly extends to simple generative tasks such as short-term forecasting.
}
\label{tab:forecasting}
\footnotesize
\setlength{\tabcolsep}{0.5em}
\begin{tabular}{lccccccccc}
    \toprule
    \textbf{Encoder} & \textbf{ETTh1} & \textbf{ETTh2} & \textbf{ETTm1} & \textbf{ETTm2} & \textbf{Weather} & \textbf{Electricity} & \textbf{Illness} & \textbf{Traffic} \\
    \midrule
    \multicolumn{4}{l}{\footnotesize\textit{\textcolor{lightgray}{Simple forecasting models}}} \\
    \texttt{N-BEATS} & 0.399 & 0.327 & 0.318 & 0.197 & 0.152 & 0.131 & 4.539 & 0.375 \\
    \texttt{DLinear} & 0.375 & 0.289 & 0.299 & 0.167 & 0.176 & 0.140 & 2.215 & 0.410 \\
    \texttt{PatchTST} & 0.370 & 0.274 & 0.293 & 0.166 & 0.149 & 0.129 & \textbf{1.319} & 0.360 \\
    \multicolumn{4}{l}{\footnotesize\textit{\textcolor{lightgray}{Specialised forecasting models}}} \\
    \texttt{TTM} & 0.363 & \underline{0.262} & \underline{0.283} & \textbf{0.158} & 0.149 & 0.128 & 2.682 & \underline{0.352} \\
    \texttt{GPT4TS} & 0.376 & 0.285 & 0.292 & 0.173 & 0.162 & 0.139 & \underline{2.063} & 0.388 \\
    \texttt{ROSE} & \underline{0.362} & 0.283 & 0.291 & 0.164 & 0.159 & 0.138 & 2.814 & 0.377 \\
    \texttt{MOIRAI} & 0.375 & 0.277 & 0.335 & 0.189 & 0.167 & 0.152 & 2.923 & 0.395 \\
    \texttt{Chronos} & 0.427 & 0.303 & 0.391 & 0.194 & 0.183 & 0.165 & 3.142 & 0.459 \\
    \texttt{Timer-XL} & 0.369 & 0.278 & 0.290 & 0.175 & 0.157 & \underline{0.127} & 2.835 & \textbf{0.340} \\
    \texttt{Time-MoE} & \textbf{0.344} & 0.275 & \textbf{0.271} & 0.179 & 0.159 & 0.147 & 2.673 & 0.367 \\
    \texttt{DTAF} & 0.382 & 0.289 & 0.306 & 0.175 & \underline{0.149} & 0.143 & 2.163 & 0.386 \\
    \midrule
    \multicolumn{4}{l}{\footnotesize\textit{\textcolor{lightgray}{General-purpose encoders}}} \\
    \texttt{MOMENT-386M} & 0.387 & 0.288 & 0.293 & 0.170 & 0.154 & 0.136 & 2.728 & 0.391 \\
    \texttt{UniTS-8.2M} & 0.394 & 0.361 & 0.369 & 0.259 & 0.216 & 0.152 & 2.472 & 0.421 \\
    \rowcolor{highlight}
    \texttt{OTIS-7.1M} & 0.419 & \textbf{0.247} & 0.337 & \underline{0.162} & \textbf{0.139} & \textbf{0.125} & 2.643 & 0.387 \\
    \bottomrule
\end{tabular}
\end{table}

\paragraph{Generative Capabilities}
We further explore whether \texttt{OTIS} features can even be utilised for simple generative tasks.
By reusing the lightweight $1.5\,$M decoder from pre-training, we evaluate performance on standard short-term forecasting (Tab.~\ref{tab:forecasting}).
\texttt{OTIS} outperforms \texttt{UniTS} and \texttt{MOMENT} on $5$ of $8$ datasets, suggesting it encodes \textit{temporal patterns} rich enough to predict short future horizons from past observations (App. \ref{sec:app_forecast_vis}).

\begin{wrapfigure}[17]{r}
{0.5\linewidth}
    \centering
    \vspace{-15pt}
    \includegraphics[width=1.0\linewidth]{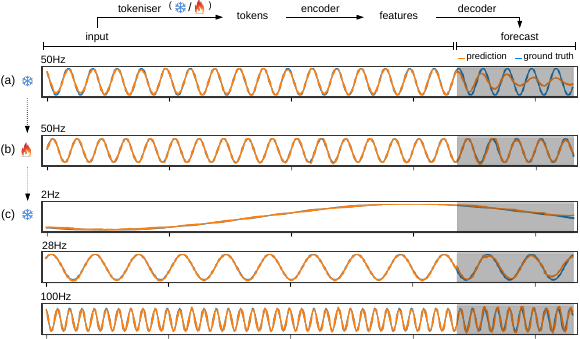}
    \caption{
    \textbf{\texttt{OTIS} features encode temporal patterns.}
    \texttt{OTIS} and the decoder are \textit{frozen} after pre-training.
    (a)~The decoder predicts the input frequency, confirming \texttt{OTIS} extracts relevant temporal features.
    (b)~Fine-tuning \textit{solely} the variate embedding mitigates mean collapse over longer horizons without updating any encoder and decoder weights.
    (c)~The learned $50\,$Hz semantics transfer to \textit{unseen} frequencies, demonstrating \texttt{OTIS} generalises beyond the training distribution.
    }
    \label{fig:temporal_dynamics}
\end{wrapfigure}

To isolate this capability, we perform controlled experiments on sine waves while keeping both \texttt{OTIS} and the decoder completely frozen after pre-training (Fig.~\ref{fig:temporal_dynamics}).
We feed \texttt{OTIS} a $50\,$Hz sine wave over time steps $t \in [0, T_\text{input}]$ and pass the extracted features to the decoder to forecast $t \in [{T_\text{input}+1}, T_\text{forecast}]$.
The resulting predictions reveal that the frozen decoder faithfully reproduces the input frequency (Fig.~\ref{fig:temporal_dynamics}a), confirming features produced by \texttt{OTIS} include temporal patterns useful for short-term forecasting. 
While predicted time steps suffer from gradual mean-collapse the more they are in the future, this can be simply mitigated by fine-tuning \textit{solely} the learnable variate embedding in the tokeniser (Fig.~\ref{fig:temporal_dynamics}b).
This noticeably improves the quality of the extracted feature, enabling the frozen decoder to generalise and predict frequencies of \textit{unseen} sine waves (Fig.~\ref{fig:temporal_dynamics}c).
Together, these results show that features extracted by \texttt{OTIS} (i) encode temporal patterns suitable for generative tasks, and (ii) can be adapted to new domains simply by fine-tuning the variate embeddings ($V_S\times D$ parameters) on a single target sample.

\subsection{\texttt{OTIS} Learns a Well-Structured Representation Space}
\label{sec:ts_understanding}
To understand \textit{why} \texttt{OTIS} features generalise across various tasks and domains, we investigate the representation space learned during pre-training.

\begin{figure}[!t]
    \centering
    \newcommand{\figheight}{4cm}

    \begin{subfigure}[t]{0.32\textwidth}
        \centering
        \includegraphics[height=\figheight]{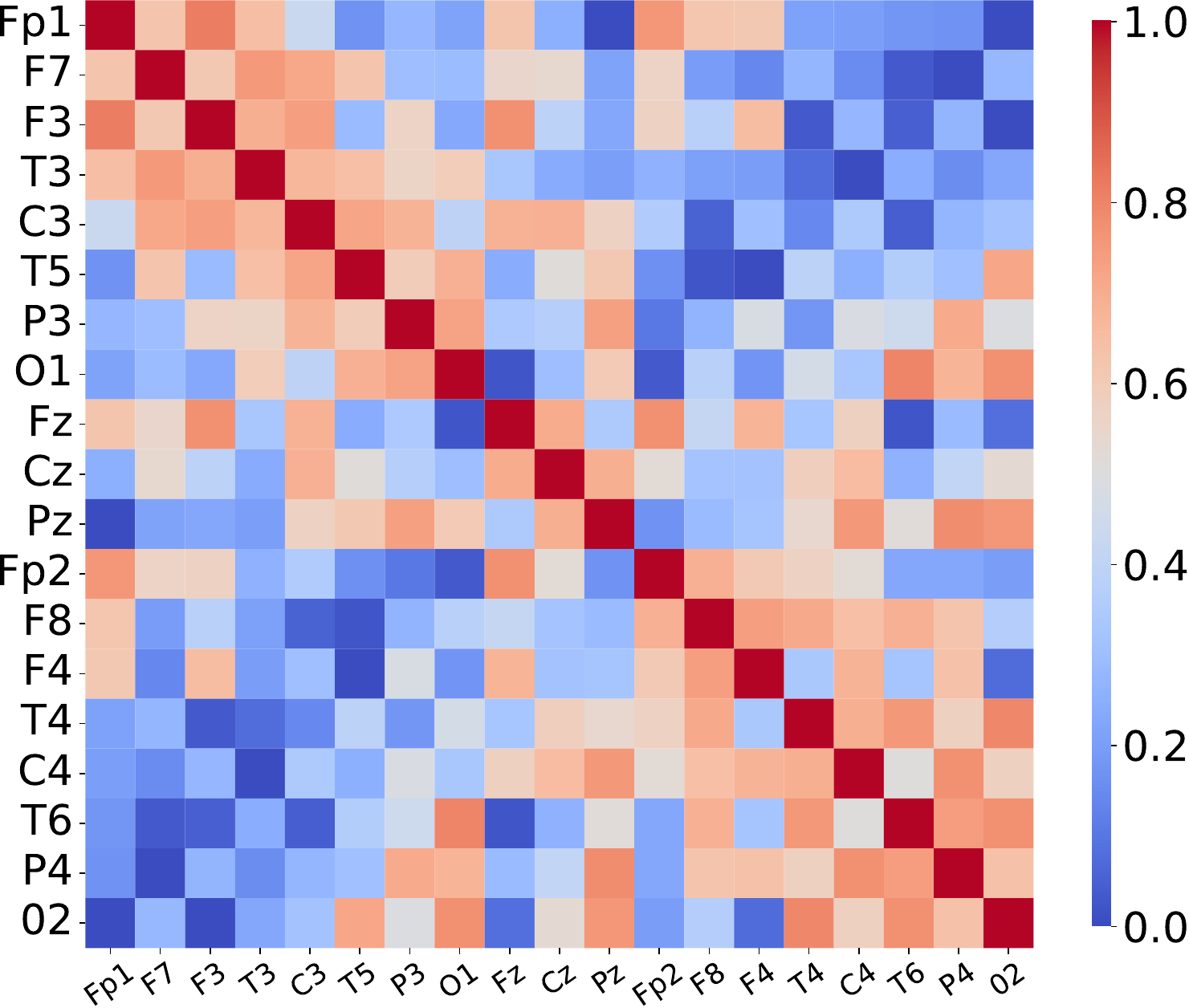}
        \caption{EEG.}
        \label{fig:eeg}
    \end{subfigure}
    \hfill
    \begin{subfigure}[t]{0.32\textwidth}
        \centering
        \includegraphics[height=\figheight]{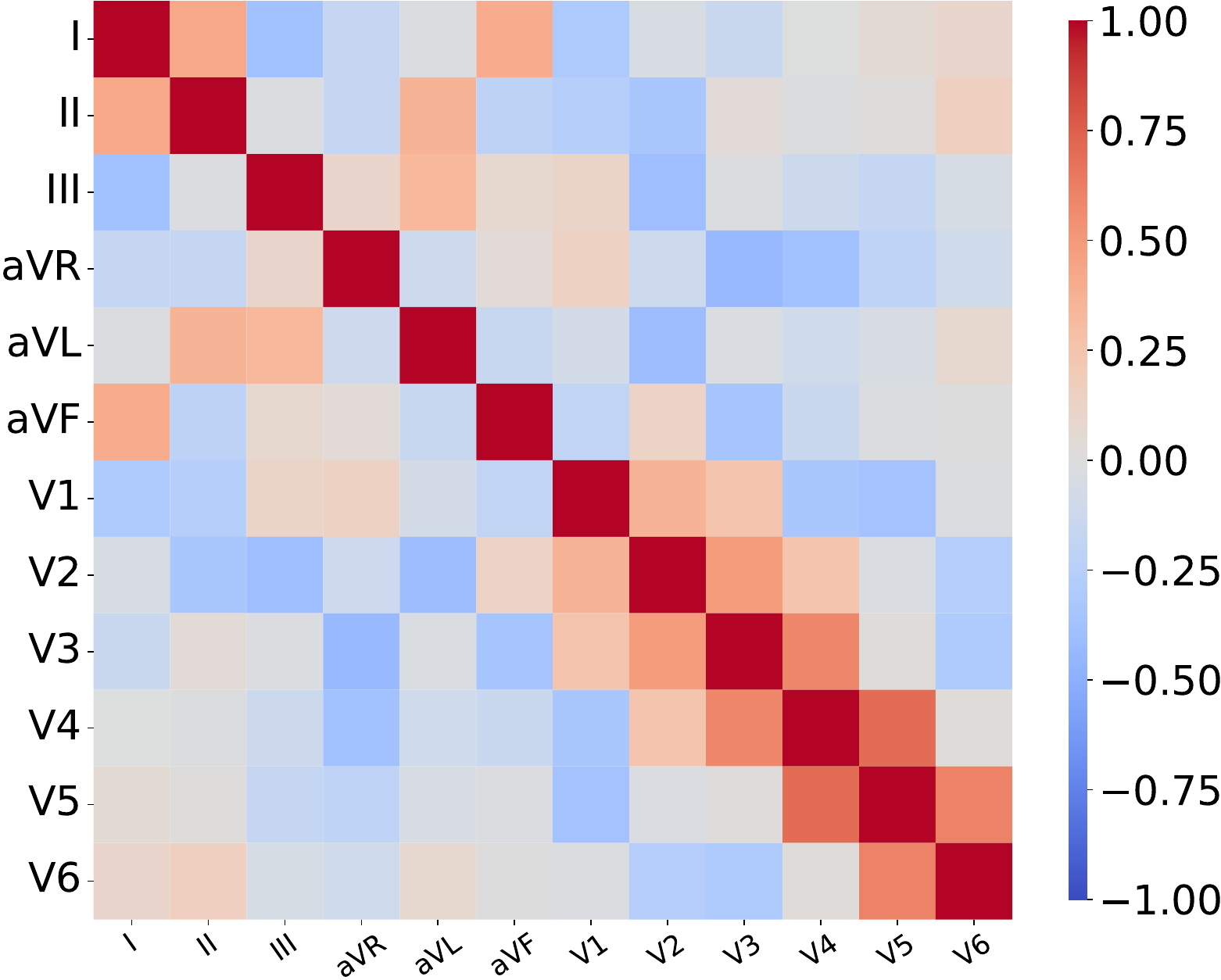}
        \caption{ECG.}
        \label{fig:ecg}
    \end{subfigure}
    \hfill
    \begin{subfigure}[t]{0.32\textwidth}
        \centering
        \includegraphics[height=\figheight]{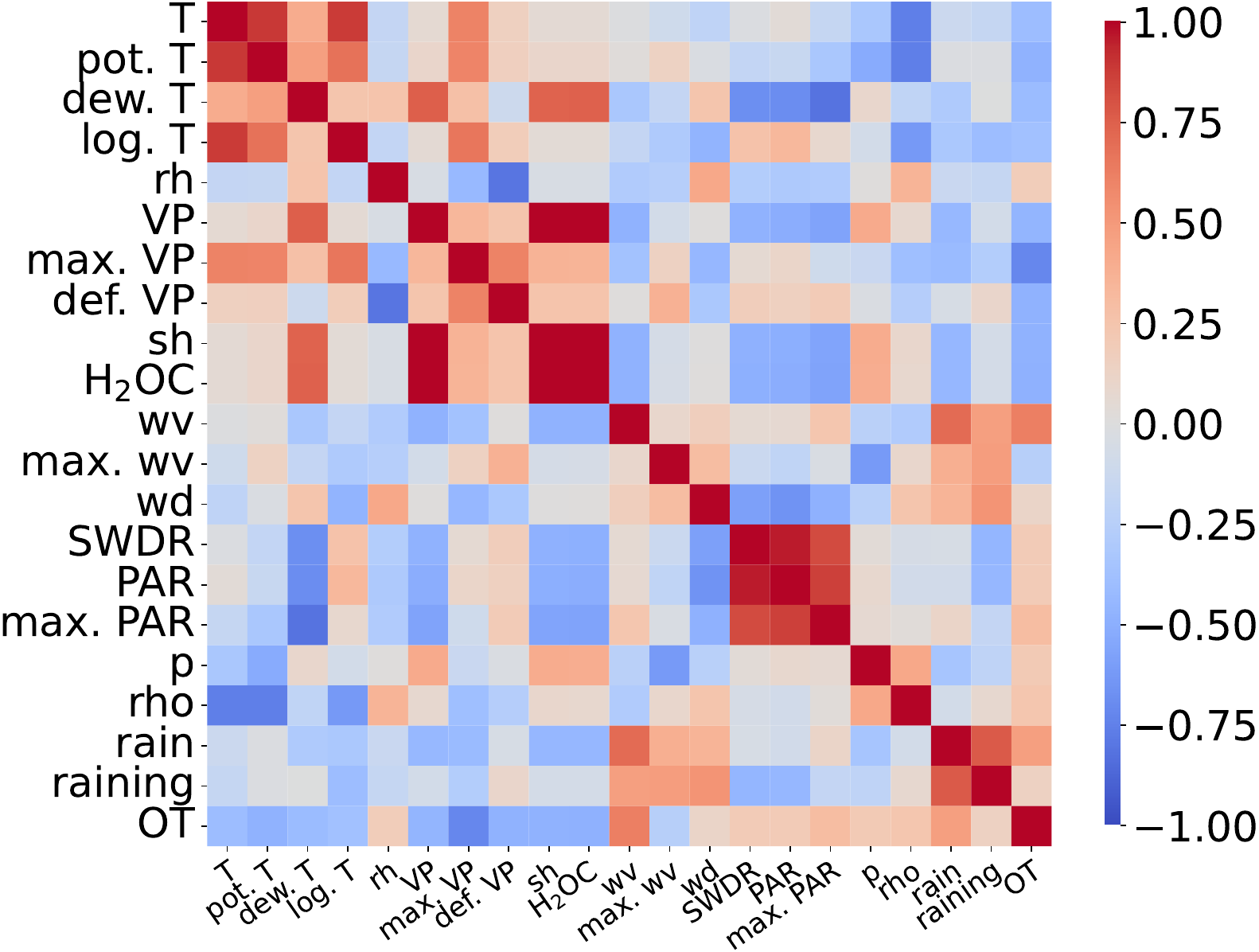}
        \caption{Weather.}
        \label{fig:weather}
    \end{subfigure}

    \caption{
    \textbf{Isolating multi-domain data heterogeneity.}
    Cosine similarities within variate embeddings $\mathbf{e}^\mathcal{V}_{S} \in \mathbb{R}^{V_S \times D}$. 
    Domain semantics of (a) EEG, (b) ECG, and (c) Weather are correctly isolated by the tokeniser.
    }
    \label{fig:cos_sim}
\end{figure}

\paragraph{Isolating Multi-Domain Data Heterogeneity}
\label{sec:semantic_shifts}
Different domains have unique semantics especially across variates.
If these semantical differences are not isolated, they become conflicting learning signals for the encoder.
We find that learnable variate embeddings successfully address this heterogeneity (Fig.~\ref{fig:cos_sim}).

\textit{Neurology.}
EEG electrodes are spatially arranged according to the standard $10$-$20$ system \citep{Homan1987}.
The learned variate embeddings successfully recover this underlying structure (Fig.~\ref{fig:eeg}).
We observe strong similarity among electrodes within the left hemisphere (Fp1 to O1), right hemisphere (Fp2 to O2), and the sagittal plane (Fz, Cz, Pz).
Conversely, low similarity corresponds to electrodes with distant separation.

\textit{Cardiology.}
Similarly, the learned variate embeddings for $12$-lead ECG reflect the standard acquisition protocol (Fig.~\ref{fig:ecg}).
We find high similarity among the precordial leads V$1$--V$6$ \citep{Wilson1934}, which physically span the rib cage.
Furthermore, the limb leads I, II, and III \citep{Einthoven1902} exhibit similarities consistent with the Einthoven triangle \citep{Kligfield2007}.

\textit{Meteorology.}
The learned variate embeddings for climatological indicators cluster by thermodynamic category: temperature (T), humidity (rh, VP, sh, H$_2$OC), wind (wv, wd), radiation (SWDR, PAR), pressure (p, rho), and precipitation (rain, raining) (Fig.~\ref{fig:weather}).
Correlations across categories are also captured, such as the thermodynamic law $\text{def. VP} \propto (1 - \text{rh})$ \citep{Shamshiri2018} between vapor pressure deficit (def. VP in mbar) and relative humidity (rh in $\%$), and the more intuitive inverse relationship between precipitation (raining in seconds) and solar radiation (SWDR in W/m$^2$).

\begin{figure}[!t]
    \centering
    \begin{subfigure}[t]{0.19\textwidth}
        \centering
        \includegraphics[width=1.0\linewidth]{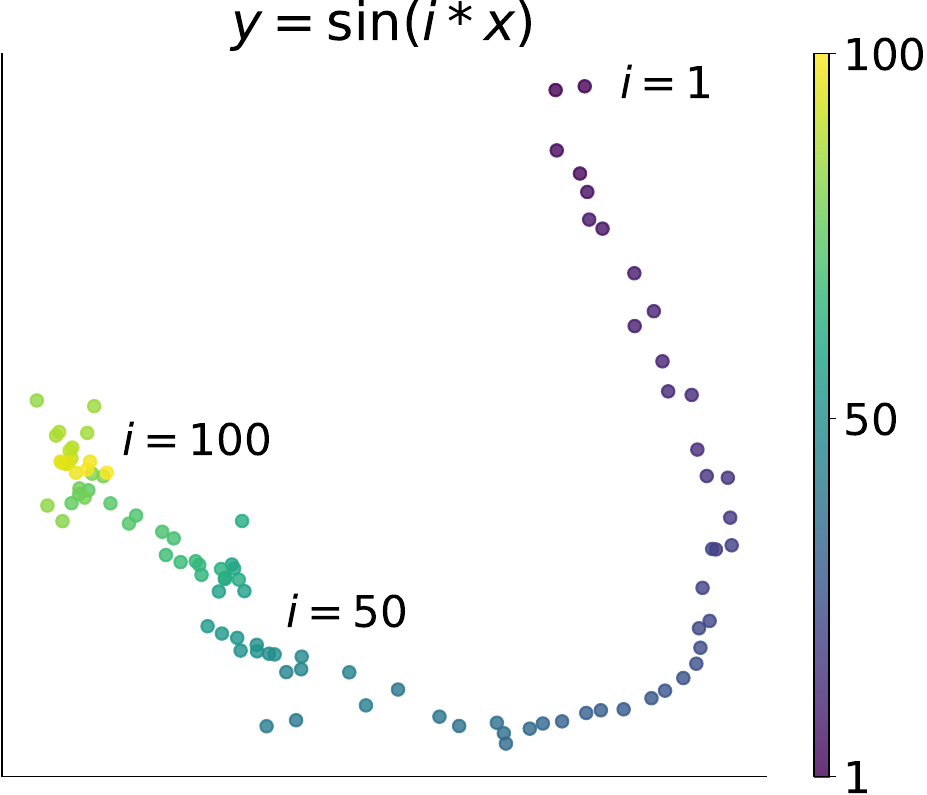}
        \caption{Frequency.}
        \label{fig:zero_frequency}
    \end{subfigure}
    \begin{subfigure}[t]{0.19\textwidth}
        \centering
        \includegraphics[width=0.99\linewidth]{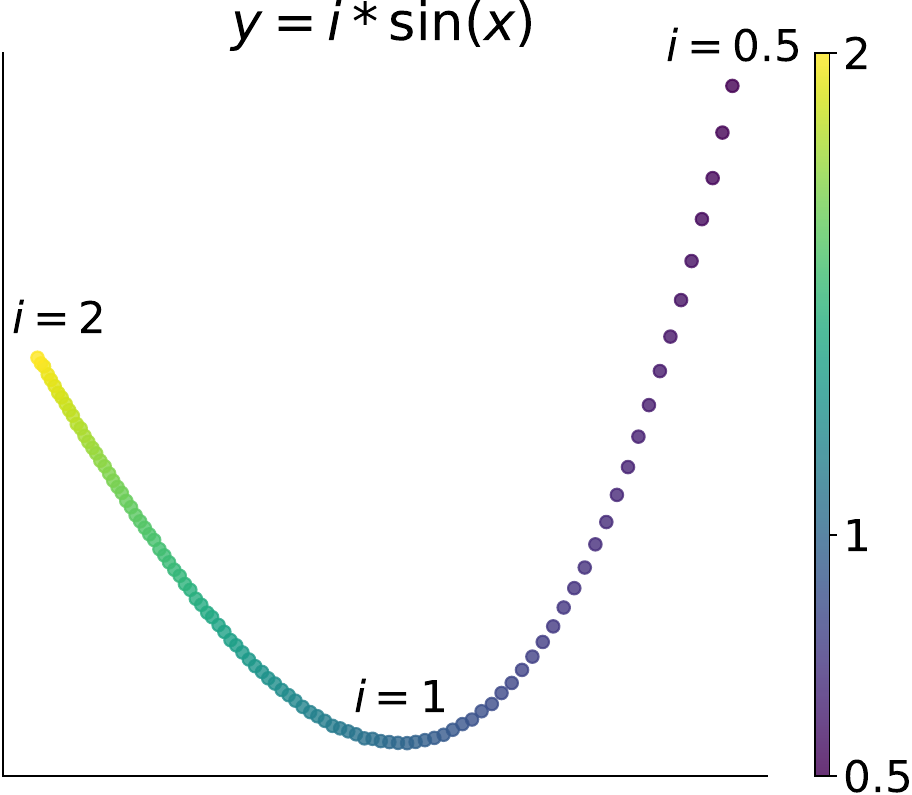}
        \caption{Amplitude.}
        \label{fig:zero_amplitude}
    \end{subfigure}
    \begin{subfigure}[t]{0.19\textwidth}
        \centering
        \includegraphics[width=0.94\linewidth]{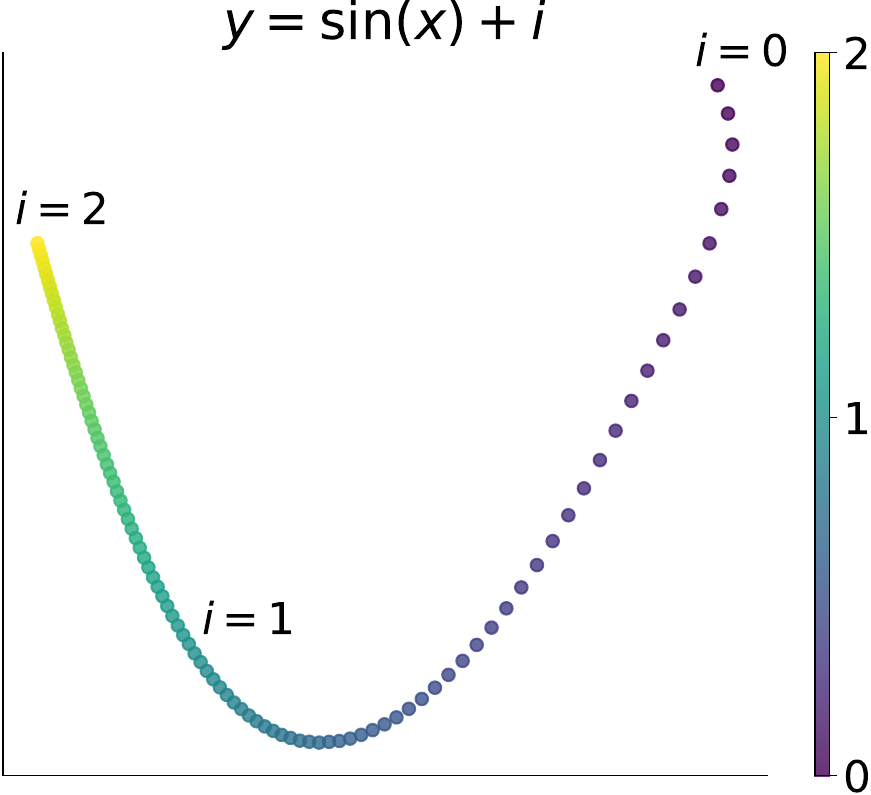}
        \caption{Offset.}
        \label{fig:zero_offset}
    \end{subfigure}
    \begin{subfigure}[t]{0.19\textwidth}
        \centering
        \includegraphics[width=0.98\linewidth]{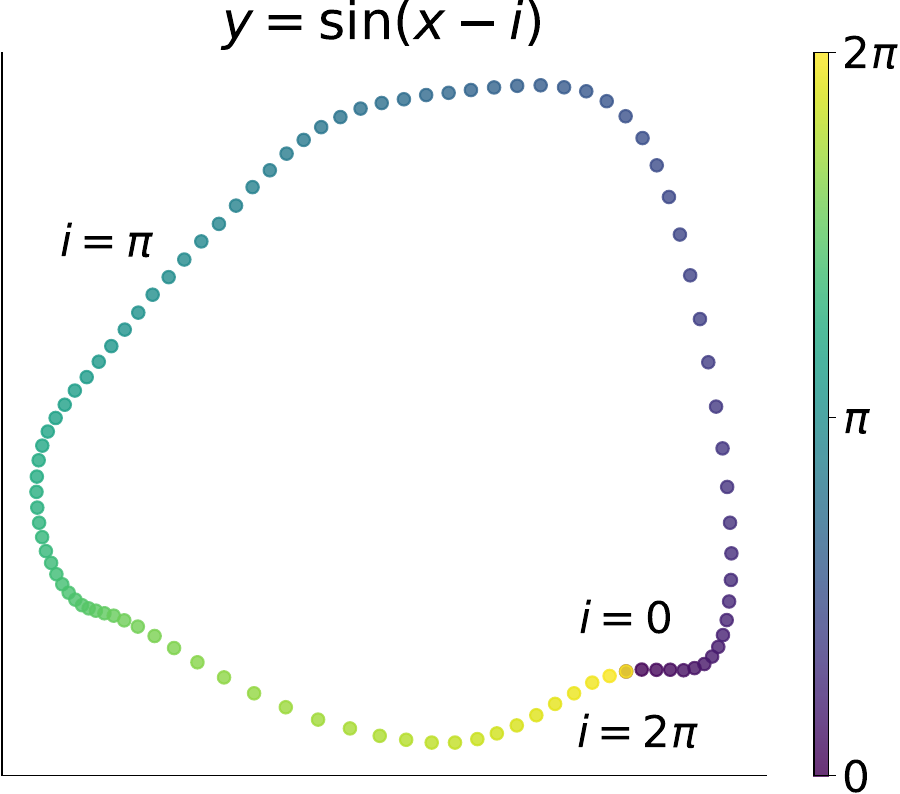}
        \caption{Phase.}
        \label{fig:zero_phase}
    \end{subfigure}
    \begin{subfigure}[t]{0.19\textwidth}
        \centering
        \includegraphics[width=1.01\linewidth]{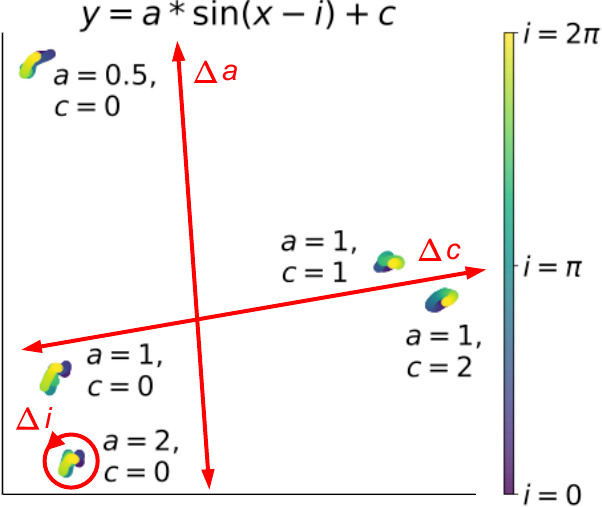}
        \caption{Disentanglement.}
        \label{fig:zero_composition}
    \end{subfigure}
    \caption{
    \textbf{Learning a consistent representation space.}
    Principal component analysis of \textit{zero-shot} sine wave representations.
    (a-c) Continuous properties map to logarithmic manifolds.
    (d) Phase maps to a closed circle.
    (e) Simultaneous changes in properties are orthogonally disentangled in the representation space.
    }
    \label{fig:general_features}
\end{figure}

\paragraph{Learning a Consistent Representation Space}
\label{sec:general_features}
With multi-domain data heterogeneity addressed by the tokeniser, the encoder is free to learn a well-structured, consistent representation space.
To validate this, we freeze \texttt{OTIS} after pre-training and run controlled experiments on sine waves with distinct properties (Fig.~\ref{fig:general_features}).

\textit{Manifold Geometry.}
To explore the structure of the learned representation space, we probe \texttt{OTIS} using sine waves with varying frequency, amplitude, phase, and offset.
We find that the representation space exhibits a well-defined geometry for these time series properties.
For frequency (Fig.~\ref{fig:zero_frequency}), the representations follow a logarithmic mapping, where high frequencies (e.g. $50$--$100\,$Hz) are clustered more tightly than low frequencies (e.g. $1$--$50\,$Hz).
This aligns with intuition: an absolute increase of $1\,$Hz from $5$ to $6\,$Hz reflects a relative change of $20\,\%$, whereas an increase from $50$ to $51\,$Hz reflects only a change of $2\,\%$.
Similarly, amplitude (Fig.~\ref{fig:zero_amplitude}) and offset (Fig.~\ref{fig:zero_offset}) follow logarithmic mappings.
Most notably, the mapping of phase (Fig.~\ref{fig:zero_phase}) forms a closed circle.
The representation for $\phi=0$ is identical to $\phi=2\pi$ and opposed to $\phi=\pi$, where $\phi \in [0, 2\pi]$ denotes the phase.
This topological closure correctly mirrors the unit circle $e^{i \phi} = \text{cos}(\phi) + i \cdot \text{sin}(\phi)$.
Consequently, this continuous and topologically consistent representation space proves the encoder learns the \textit{actual} structure of these properties rather than memorising distributions of the pre-training data.
This empowers \texttt{OTIS} to generalise to unseen time series, making it a robust general-purpose encoder despite its tiny size.

\textit{Orthogonal Disentanglement.}
Crucially, when multiple properties vary simultaneously, the encoder disentangles them into orthogonal subspaces of the representation space (Fig.~\ref{fig:zero_composition}).
Changes in amplitude do not distort the representation with respect to phase or offset, suggesting these properties are treated as independent variables.
This orthogonality simplifies downstream adaptation of \texttt{OTIS}, allowing task-specific heads to isolate relevant signal properties (e.g. amplitude) without interference from irrelevant variations (e.g. phase shifts).

\subsection{\texttt{OTIS} Benefits from Targeted Inductive Bias}
\label{sec:ablation}
\begin{table}[t]
\centering
\caption{
\textbf{Inductive bias is essential.}
The average performance gain (PG) contributed by each component is evaluated in a leave-one-out ablation.
Components are mutually reinforcing, with the removal of any single bias leading to suboptimal performance.
A structure-aware objective supplies the largest individual gains.
}
\label{tab:ablation}
\footnotesize
\setlength{\tabcolsep}{1.0em}
\begin{tabular}{lllllll}
    \toprule
    \multirow{2}{*}{{\textbf{Encoder}}} & \multicolumn{2}{l}{{\textbf{Classification}}} & \multicolumn{2}{l}{\textbf{{Regression}}} & \multicolumn{2}{l}{\textbf{{Forecasting}}} \\

    {} & ACC $\uparrow$ & PG & $R^2$ $\uparrow$ & PG & MSE $\downarrow$ & PG \\

    \midrule
    \rowcolor{Gainsboro!60}
    \texttt{OTIS} & $85.7$ &  & $0.442$ &  & $0.976$ & \\

    \texttt{ w/o domain-aware tokenisation} & $84.5$ & \footnotesize{\textcolor{teal}{$\uparrow 1.42\,\%$}} & $0.441$ & \footnotesize{\textcolor{teal}{$\uparrow 0.23\,\%$}} & $1.075$ & \footnotesize{\textcolor{teal}{$\uparrow 9.21\,\%$}} \\

    \texttt{ w/o dual masking} & $84.1$ & \footnotesize{\textcolor{teal}{$\uparrow 1.90\,\%$}} & $0.443$ & \footnotesize{\textcolor{red!70}{$\downarrow 0.23\,\%$}} & $0.995$ & \footnotesize{\textcolor{teal}{$\uparrow 1.91\,\%$}} \\

    \texttt{ w/o normalised cross-correlation loss} & $83.5$ & \footnotesize{\textcolor{teal}{$\uparrow 2.63\,\%$}} & $0.438$ & \footnotesize{\textcolor{teal}{$\uparrow 0.91\,\%$}} & $1.161$ & \footnotesize{\textcolor{teal}{$\uparrow 15.93\,\%$}} \\

    \texttt{ w/o pre-training} & $75.1$ & \footnotesize{\textcolor{teal}{$\uparrow 14.11\,\%$}} & $0.322$ & \footnotesize{\textcolor{teal}{$\uparrow 37.27\,\%$}} & $2.077$ & \footnotesize{\textcolor{teal}{$\uparrow 53.00\,\%$}} \\
    \bottomrule
\end{tabular}
\end{table}
We conduct a comprehensive ablation study to quantify the impact of the three proposed inductive biases: (1) the domain-aware tokeniser, (2) the dual masking strategy, and (3) the structure-aware objective.
First, replacing the domain-aware tokeniser with a domain-agnostic one (i.e. universal positional embeddings shared across all domains) consistently degrades performance across all tasks.
This confirms that explicitly resolving semantic heterogeneity by design is a prerequisite for effective representation learning in multi-domain settings.
Second, reverting from dual to purely random masking harms performance, particularly in forecasting.
This validates our hypothesis that forcing the encoder to predict future horizons during pre-training is critical for learning temporal patterns.
Third, removing the structure-aware NCC loss term from the training objective causes the most significant performance decline.
This underscores that relying solely on local point-wise reconstruction via MSE is insufficient; explicitly enforcing global structural alignment is essential for learning robust time series representations.
Overall, these ablations demonstrate that inductive biases are mutually reinforcing and their interplay is essential to maximise the feature quality of a tiny encoder.

\subsection{\texttt{OTIS} Scales with Data}
\label{sec:scaling}
\begin{figure}[t]
    \centering
    \newcommand{\figheight}{2.9cm}
    \begin{subfigure}[t]{0.325\linewidth}
        \centering
        \includegraphics[height=\figheight]{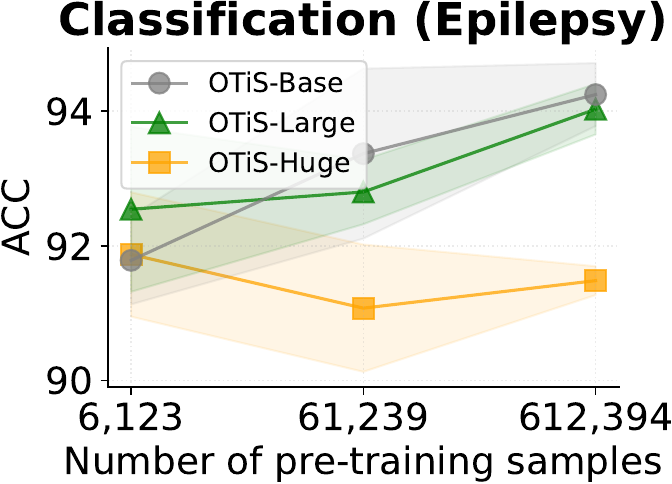}
    \end{subfigure}
    \begin{subfigure}[t]{0.325\linewidth}
        \centering
        \includegraphics[height=\figheight]{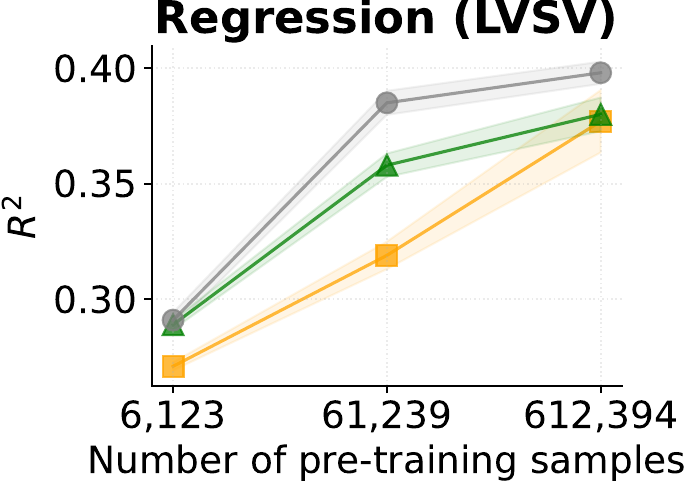}
    \end{subfigure}
    \begin{subfigure}[t]{0.325\linewidth}
        \centering
        \includegraphics[height=\figheight]{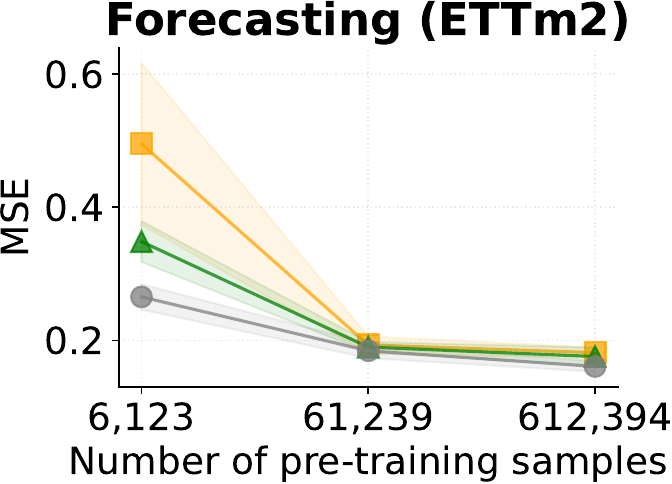}
    \end{subfigure}
    \caption{
    \textbf{Scaling is not trivial.}
    Downstream performance scales reliably with the size of the pre-training corpus ($1\,\%$, $10\,\%$, $100\,\%$), but saturates with encoder size: Large ($40\,$M) and Huge ($116\,$M) variants offer little advantage over the $7.1\,$M base.
    Shaded regions indicate standard deviation across five seeds.
    }
    \label{fig:low_data}
\end{figure}
Having shown that our simple pre-training recipe enables standard masked data modelling to produce robust features, we now examine its scalability.
To this end, we analyse the scaling behaviour of \texttt{OTIS} with respect to both dataset and encoder size (Fig.~\ref{fig:low_data}).
We train variants using subsamples of the pre-training data ($1\,\%$, $10\,\%$, $100\,\%$) and introduce two larger configurations of our encoder: Large ($40\,$M) and Huge ($116\,$M) (App.~\ref{sec:app_exp_details}).
Our analysis reveals two distinct trends.
First, \textit{performance scales reliably with data}.
We observe consistent downstream gains as the pre-training corpus grows, indicating that the encoder effectively leverages larger datasets to refine its representation space.
Second, \textit{performance saturates with encoder size}.
We find that Large and Huge encoder variants offer almost no performance advantage over the tiny $7.1\,$M variant, which is in line with previous studies on time series analysis \citep{Woo2024, Goswami2024, Liu2026}.
This plateau suggests that for time series, which often exhibit redundancy especially in multi-variate settings, producing arbitrarily large, powerful encoders is not as simple as increasing parameter counts, even when abundant pre-training data is available \citep{Liu2026}.
Instead, extensive engineering might be required, much like in natural language processing and computer vision, where naive scaling is often hindered by training instabilities and feature collapse \citep{Simeoni2025, Olmo2025}.

\section{Discussion \& Conclusion}
\label{sec:conclusion}
In this work, we introduce \texttt{OTIS} to challenge the prevailing assumption that powerful general-purpose encoder require massive scaling.
By strategically injecting inductive biases through (1) a \textit{domain-aware tokeniser}, (2) a \textit{dual masking strategy}, and (3) a \textit{structure-aware objective}, we show that a tiny $7.1\,$M encoder can match or even exceed state-of-the-art encoders orders of magnitude larger, while saving $90\,\%$ memory, $98\,\%$ energy, and $97\,\%$ latency.
Our extensive evaluation yields three insights.
First, \texttt{OTIS} learns a topologically consistent representation space that enables robust generalisation across diverse tasks and domains, ranging from industrial fault detection to clinical diagnostics.
Second, the proposed inductive biases are mutually reinforcing; removing any single component notably degrades performance, confirming their collective necessity for learning high-quality time series features.
Third, while performance scales reliably with data size, simply scaling parameter counts of time series encoders yields diminishing returns.
Together, these findings demonstrate that targeted inductive bias offers a promising alternative to the development of general-purpose encoders orthogonal to traditional scaling.
By delivering competitive performance alongside highly hardware deployability, \texttt{OTIS} marks a distinct step towards democratising access to high-quality time series features on resource-constrained systems such as wearables and industrial edge sensors.

\subsubsection*{Limitations}
While \texttt{OTIS} matches state-of-the-art performance with a fraction of the parameters, our study surfaces several open challenges.
First, performance scales reliably with data but saturates with encoder size: Large ($40\,$M) and Huge ($116\,$M) variants offer little advantage over the $7.1\,$M base, indicating that scaling time series encoder size requires further engineering and methodological advances.
Second, unlike in NLP and CV, general-purpose time series encoders still rely on manually curated data; automated pipelines to filter and categorise time series from the web are a natural next step.
Third, our evaluation is restricted to continuous time series, leaving the integration and evaluation of irregularly sampled time series to future work.

\subsubsection*{Statement of Broader Impact}
This work aims to advance the field of Time Series Analysis.
By developing tiny general-purpose time series encoders that can in future be deployed on resource-constrained devices like wearables and industrial sensors, our work has potential positive societal impacts in democratising access to high-quality time series features.
However, as with any general-purpose backbone, there are risks associated with deployment in safety-critical domains without sufficient validation.
We believe further research into the robustness and interpretability of such tiny encoders is necessary before deployment, particularly in clinical and industrial routine.

\bibliography{otis}
\bibliographystyle{tmlr}

\newpage
\appendix
\section{Pre-Training Corpus Details}
\label{sec:app_ts_corpus}
In this section, we present an overview of our large and diverse pre-training corpus, as summarised in Table \ref{tab:train_datasets}.
The corpus consists of publicly available data spanning eight domains, with a total of $640,187$ samples and $11$ billion time points.
In the following, we provide a detailed breakdown of the domains and the datasets they encompass.
Note that we apply variate-wise standard normalisation to the datasets unless otherwise specified.

\paragraph{ECG}
The MIMIC-IV-ECG dataset \citep{Gow2023} contains diagnostic $10$-second, $12$-lead ECG recordings sampled at a frequency of $500\,$Hz.
While the entire dataset comprises $800,035$ samples, we include only the first half of the recordings available in the database, preventing the ECG data from predominating in the pre-training corpus.
To remove the baseline drift from the ECG data, we use asymmetric least square smoothing \citep{Zhang2010}.
We apply group-wise standard normalisation to the Einthoven, Goldberger, and Wilson leads, respectively.

\paragraph{Temperature}
The Deutscher Wetterdienst (DWD) dataset \citep{DWD2024} contains hourly air temperature measurements from $629$ weather stations across Germany.
Since the recording length varies significantly, ranging from $763$ to $1,148,290$ hours per station, we split the data into chunks of $720$ hours (approximately one month).

\paragraph{Audio}
The AudioSet dataset \citep{Gemmeke2017} contains $10$-second YouTube clips for audio classification, featuring $527$ types of audio events that are weakly annotated for each clip.
The full training set includes a class-wise balanced subset (AudioSet-$20$K, $22,176$ clips) and an unbalanced (AudioSet-$2$M $2,042,985$ clips) set.
For our pre-training corpus, we use the balanced AudioSet-$20$K, which contains $3,491$ mono and $16,123$ stereo recordings, all sampled at $44.1\,$kHz.

\paragraph{Electromechanics}
The FD-A dataset \citep{Lessmeier2016} collects vibration signals from rolling bearings in a mechanical system for fault detection purposes.
Each sample consists of $5,120$ timestamps, indicating one of three mechanical device states.
The FD-B dataset contains bearing-vibration signals from the same machine, but under strictly different working conditions (rotational speed, load, torque, and radial force).

\paragraph{EEG}
The TDBrain dataset \citep{VanDijk2022} includes raw resting-state EEG data from $1,274$ psychiatric patients aged $5$ to $89$, collected between $2001$ and $2021$.
The dataset covers a range of conditions, including Major Depressive Disorder ($426$ patients), Attention Deficit Hyperactivity Disorder ($271$ patients), Subjective Memory Complaints ($119$ patients), and Obsessive-Compulsive Disorder ($75$ patients).
The data was recorded at $500\,$Hz using $26$ electrode EEG.

The SEED dataset \citep{Zheng2015} contains EEG data recorded under three emotional states: positive, neutral, and negative.
It comprises EEG data from $15$ subjects, with each subject participating in experiments twice, several days apart.
The data is sampled at $200\,$Hz and recorded using $62$ electrode EEG.

For simplicity, we only consider the $19$ electrodes common to both datasets, i.e. the electrodes that correspond to the $10$-$20$ electrode international system \citep{Homan1987}.

\paragraph{Banking}
The NN$5$ competition dataset \citep{Taieb2012} consists of daily cash withdrawals observed at $111$ randomly selected automated teller machines across various locations in England.

\paragraph{Economics}
The FRED-MD dataset \citep{McCracken2016} contains $107$ monthly time series showing a set of macro-economic indicators from the Federal Reserve Bank of St Louis.
The data was extracted from the FRED-MD database.

The Exchange dataset \citep{Lai2018} records the daily exchange rates of eight different nations, including Australia, Great Britain, Canada, Switzerland, China, Japan, New Zealand, and Singapore, ranging from $1990$ to $2016$.

\begin{table}[!t]
\caption{
The pre-training data covers domains in healthcare, engineering, natural sciences, and finance.}
\label{tab:train_datasets}
\centering
\small
\setlength{\tabcolsep}{1.05em}
\begin{tabular}{llrrrrr}
    \toprule
    \textbf{Domain} & \textbf{Name} & \textbf{Samples} & \textbf{Variates} & \textbf{Time points} & \textbf{Frequency} \\
    \midrule
    ECG   & MIMIC-IV-ECG & $400,000$ & $12$ & $5,000$   & $500\,$Hz \\
    Temperature  & DWD & $203,340$ & $1$  & $720$     & $278\,\mu$Hz \\
    Audio (stereo) & AudioSet-$20$K & $16,123$  &  $2$ & $441,000$ & $44.1\,$kHz \\
    Audio (mono) & AudioSet-$20$K & $3,491$  &  $1$ & $441,000$ & $44.1\,$kHz \\
    Electromechanics & FD-A & $13,640$ & $1$ & $5,120$ & $64\,$kHz \\
    \multirow{2}{*}{EEG} & TDBrain & $2,692$   & $19$ & $60,000$  & $500\,$Hz \\
    {}   & SEED & $675$ & $19$  & $37,000$  & $200\,$Hz \\
    Banking & NN$5$ & $111$ & $1$ & $971$ & $12\,\mu$Hz \\
    \multirow{2}{*}{Economics} & FRED-MD & $107$ & $1$ & $728$ & $386\,$nHz \\
    {} & Exchange & $8$ & $1$ & $7,588$ & $12\,\mu$Hz \\
    \midrule
    & & $\textbf{640,187}$ & & $\textbf{11,052,756,981}$ & \\
    \bottomrule
\end{tabular}
\end{table}

\section{Computational Details}
\label{sec:app_comp_cost}
We provide an overview of the computational resources used for pre-training in Table \ref{tab:app_pt_cost}.
Details on pre-training the larger encoder configurations, Large and Huge, described in Section \ref{sec:app_encoder_variants}, are also included.
\begin{table}[!ht]
\caption{Computational resources used for pre-training.}
\label{tab:app_pt_cost}
\centering
\normalsize
\setlength{\tabcolsep}{0.575em}
\begin{tabular}{lrccccc}
    \toprule
    \multirow{2.5}{*}{\textbf{Encoder}} & \multirow{2.5}{*}{\textbf{Parameters}} & \multirow{2.5}{*}{\makecell{\textbf{Power}\\\textbf{consumption}}} & \multirow{2.5}{*}{\makecell{\textbf{CPU}\\\textbf{count}}} & \multicolumn{3}{c}{\textbf{GPU}} \\
    \cmidrule(lr){5-7}
     & & & & \textbf{Count} & \textbf{Hours} & \textbf{Type} \\
    \midrule
    \texttt{OTIS}       & $7.1\,$M & $700\,$W* & $128$ & $4$ & $86$$^\dagger$  & NVIDIA A$100$-$80$GB \\
    \texttt{OTIS}-Large &  $40\,$M & $800\,$W* & $128$ & $4$ & $116$$^\dagger$ & NVIDIA A$100$-$80$GB \\
    \texttt{OTIS}-Huge  & $116\,$M & $960\,$W* & $128$ & $4$ & $164$$^\dagger$ & NVIDIA A$100$-$80$GB \\
    \bottomrule
    \multicolumn{7}{l}{\footnotesize{* Total power consumption across all GPUs.}} \\
    \multicolumn{7}{l}{\footnotesize{$^\dagger$ Total hours across all GPUs.}}
\end{tabular}
\end{table}

\section{Experimental Details}
\label{sec:app_exp_details}
We outline encoder configurations in Section \ref{sec:app_encoder_variants} and hyperparameter tuning in Section \ref{sec:app_hyperparams}.

\subsection{Encoder Configurations}
\label{sec:app_encoder_variants}
To explore the scaling laws with respect to the encoder size in Section \ref{sec:scaling}, we introduce two larger configurations of our encoder $f(\cdot)$, Large and Huge, as summarised in Table \ref{tab:app_encoder_variants}.

\begin{table}[!ht]
\caption{Details of encoder configurations.}
\label{tab:app_encoder_variants}
\centering
\normalsize
\setlength{\tabcolsep}{0.9em}
\begin{tabular}{lcccccr}
    \toprule
    \textbf{Encoder} & \textbf{Layers} & \textbf{Hidden size \textit{D}} & \textbf{MLP size} & \textbf{Heads} & \textbf{$d_{kv}$} & \textbf{Parameters} \\
    \midrule
    \texttt{OTIS}       & $12$ & $192$ &  $768$ & $3$ & $64$ &   $7.1\,$M \\
    \texttt{OTIS}-Large & $18$ & $384$ & $1536$ & $6$ & $64$ &    $40\,$M \\
    \texttt{OTIS}-Huge  & $24$ & $576$ & $2304$ & $8$ & $72$ &   $116\,$M \\
    \bottomrule
\end{tabular}
\end{table}

\subsection{Pre-Training \& Fine-Tuning Parameters}
\label{sec:app_hyperparams}
We provide the hyperparameters used for pre-training in Table \ref{tab:app_hyperparams_pre}.
Hyperparameters used to fine-tune the encoder on classification, regression, and forecasting tasks are summarised in Table \ref{tab:app_hyperparams_ft_cls}, \ref{tab:app_hyperparams_ft_reg}, and \ref{tab:app_hyperparams_ft_forecast}, respectively.
Optimal hyperparameters are found through a grid search over the learning rate ($1$e$-6$, $3$e$-6$, $...$, $1$e$-1$, $3$e$-1$), batch size ($2^x$, $x\in[2, 3, ..., 7]$), drop path ($0.0$, $0.1$, $0.2$), layer decay ($0.5$, $0.75$), weight decay ($0.0$, $0.05$, $0.1$, $0.15$, $0.2$, $0.25$), label smoothing ($0.0$, $0.1$, $0.2$), and NCC loss weight $\lambda$ ($0.0$, $0.1$, $0.2$).

\begin{table}[!t]
\caption{
Hyperparameters for \textbf{pre-training}.
Pre-training is conducted on $4$ NVIDIA A$100$-$80$GB GPUs using a cosine learning rate scheduler with a $10\,\%$ warm-up.
All configurations employ a lightweight $1.5\,$M decoder consisting of $4$ layers, a hidden size of $160$, an MLP size of $640$, and $5$ heads.
}
\label{tab:app_hyperparams_pre}
\centering
\normalsize
\setlength{\tabcolsep}{0.6em}
\begin{tabular}{lcccccc}
    \toprule
    \textbf{Encoder} & \textbf{Epochs} & \textbf{Batch size} & \textbf{Base LR} & \textbf{NCC $\lambda$} & \textbf{Mask ratio $\rho$} & \textbf{Weight decay} \\
    \midrule
    \texttt{OTIS}  & $200$ & $5120$ & $3$e-$5$ & $0.1$ & $0.75$ & $0.10$ \\
    \texttt{OTIS}-Large & $200$ & $3328$ & $1$e-$5$ & $0.1$ & $0.75$ & $0.15$ \\
    \texttt{OTIS}-Huge  & $200$ & $2880$ & $3$e-$6$ & $0.1$ & $0.75$ & $0.05$ \\
    \bottomrule
\end{tabular}
\end{table}

\begin{table}[!t]
\caption{
Hyperparameters used to fine-tune the \textbf{classification} tasks on a single NVIDIA RTX A$6000$-$48$GB GPU.
A cosine learning rate scheduler with a $10\,\%$ warm-up is applied.
}
\label{tab:app_hyperparams_ft_cls}
\centering
\normalsize
\setlength{\tabcolsep}{0.3em}
\begin{tabular}{llccccccc}
    \toprule
    \textbf{Dataset} & \textbf{Encoder} & \textbf{Epochs} & \textbf{Batch size} & \textbf{Base LR} & \textbf{Drop path} & \textbf{\makecell{\textbf{Layer}\\\textbf{decay}}} & \textbf{\makecell{\textbf{Weight}\\\textbf{decay}}} & \textbf{\makecell{\textbf{Label}\\\textbf{smoothing}}} \\
    \midrule
    \multirow{1}{*}{FD-B} & \texttt{OTIS} & $75$ & $32$ & $3$e-$4$ & $0.0$ & $0.75$ & $0.1$ & $0.1$ \\
    \midrule
    \multirow{1}{*}{Gesture} & \texttt{OTIS} & $75$ & $32$ & $3$e-$3$ & $0.2$ & $0.50$ & $0.1$ & $0.1$\\
    \midrule
    \multirow{1}{*}{EMG} & \texttt{OTIS} & $75$ & $32$ & $1$e-$3$ & $0.2$ & $0.75$ & $0.1$ & $0.2$ \\
    \midrule
    \multirow{1}{*}{Epilepsy} & \texttt{OTIS} & $75$ & $32$ & $3$e-$4$ & $0.2$ & $0.75$ & $0.0$ & $0.0$ \\
    \midrule
    \multirow{1}{*}{SleepEDF} & \texttt{OTIS} & $20$ & $4352$ & $1$e-$4$ & $0.1$ & $0.75$ & $0.2$ & $0.2$ \\
    \midrule
    \multirow{1}{*}{TUEV} & \texttt{OTIS} & $20$ & $1216$ & $3$e-$5$ & $0.2$ & $0.75$ & $0.1$ & $0.1$ \\
    \bottomrule
\end{tabular}
\end{table}

\begin{table}[!t]
\caption{Hyperparameters used to fine-tune the \textbf{regression} tasks on a single NVIDIA RTX A$6000$-$48$GB GPU. A cosine learning rate scheduler with a $10\,\%$ warm-up is applied.}
\label{tab:app_hyperparams_ft_reg}
\centering
\normalsize
\setlength{\tabcolsep}{0.52em}
\begin{tabular}{llccccccc}
    \toprule
    \textbf{Dataset} & \textbf{Encoder} & \textbf{Epochs} & \textbf{Batch size} & \textbf{Base LR} & \textbf{Drop path} & \textbf{\makecell{\textbf{Layer}\\\textbf{decay}}} & \textbf{\makecell{\textbf{Weight}\\\textbf{decay}}} \\
    \midrule
    \multirow{1}{*}{UK BioBank} & \texttt{OTIS} & $50$ & $192$ & $3$e-$4$ & $0.2$ & $0.75$ & $0.1$ \\
    \bottomrule
\end{tabular}
\end{table}

\begin{table}[!t]
\caption{Hyperparameters used to fine-tune the \textbf{forecasting} tasks on a single NVIDIA RTX A$6000$-$48$GB GPU.
A cosine learning rate scheduler with a $10\,\%$ warm-up is applied.}
\label{tab:app_hyperparams_ft_forecast}
\centering
\normalsize
\setlength{\tabcolsep}{0.75em}
\begin{tabular}{llccccc}
    \toprule
    \textbf{Dataset} & \textbf{Encoder} & \textbf{Epochs} & \textbf{Batch size} & \textbf{Base LR} & \textbf{NCC $\lambda$} & \textbf{Weight decay} \\
    \midrule
    \multirow{1}{*}{ETTh1} & \texttt{OTIS} & $1000$ & $1$ & $3$e-$2$ & $0.1$ & $0.15$\\
    \midrule
    \multirow{1}{*}{ETTh2} & \texttt{OTIS} & $1000$ & $1$ & $3$e-$2$ & $0.2$ & $0.25$ \\
    \midrule
    \multirow{1}{*}{ETTm1} & \texttt{OTIS} & $1000$ & $1$ & $1$e-$2$ & $0.2$ & $0.25$ \\
    \midrule
    \multirow{1}{*}{ETTm2} & \texttt{OTIS} & $1000$ & $1$ & $1$e-$2$ & $0.1$ & $0.25$ \\
    \midrule
    \multirow{1}{*}{Weather} & \texttt{OTIS} & $1000$ & $1$ & $1$e-$2$ & $0.2$ & $0.25$ \\
    \midrule
    \multirow{1}{*}{Electricity} & \texttt{OTIS} & $250$ & $32$ & $1$e-$3$ & $0.0$ & $0.25$ \\
    \midrule
    \multirow{1}{*}{Illness} & \texttt{OTIS} & $1000$ & $1$ & $3$e-$1$ & $0.1$ & $0.25$ \\
    \midrule
    \multirow{1}{*}{Traffic} & \texttt{OTIS} & $100$ & $64$ & $1$e-$2$ & $0.0$ & $0.05$ \\
    \bottomrule
\end{tabular}
\end{table}

\section{Benchmark Details}
\label{sec:app_benchmark_details}
Table \ref{tab:app_eval_datasets} provides an overview of all benchmarks conducted in this study.

\begin{table}[!ht]
\caption{Summary of all datasets used for the evaluation, including details on the metrics, domains, and time series properties.
}
\label{tab:app_eval_datasets}
\centering
\footnotesize
\setlength{\tabcolsep}{0.35em}
\begin{tabular}{lcccrrrr}
    \toprule
    \multirow{2}{*}{\textbf{\normalsize Task}} & \multirow{2}{*}{\textbf{\normalsize Metric}} & \multicolumn{6}{c}{\textbf{\normalsize Dataset}} \\
    \cmidrule(lr){3-8}
    {} & {} & \textbf{\normalsize Domain} & \textbf{\normalsize Name} & \textbf{\normalsize Samples} & \textbf{\normalsize Variates} & \textbf{\normalsize Time points} & \textbf{\normalsize Frequency} \\
    {} & {} & $S$ & & & $V_S$ & & \\
    \midrule
    \multirow{6}{*}{\STAB{\rotatebox[origin=c]{90}{\footnotesize{Classification}}}} & \multirow{6}{*}{ACC} & Electromechanics & FD-B \citeyear{Lessmeier2016} & $13,640$ & $1$ & $5,120$ & $64\,$kHz \\
    {} & {} & Acceleration & Gesture \citeyear{Liu2009} & $560$ & $3$ & $206$ & $100\,$Hz  \\
    {} & {} & EMG & EMG \citeyear{Goldberger2000} & $204$ & $1$ & $1,500$ & $4\,$kHz \\
    {} & {} & \multirow{3}{*}{EEG} & Epilepsy \citeyear{Andrzejak2001} & $11,500$ & $1$ & $178$ & $174\,$Hz  \\
    {} & {} & {} & SleepEDF \citeyear{Goldberger2000} & $195,479$ & $1$ & $3,000$ & $100\,$Hz \\
    {} & {} & {} & TUEV \citeyear{Obeid2016} & $112,237$ & $19$ & $1,000$ & $200\,$Hz \\
    \midrule
    \multirow{4}{*}{\STAB{\rotatebox[origin=c]{90}{\footnotesize{Regression}}}} & \multirow{4}{*}{$R^2$} & \multirow{4}{*}{ECG} & \multirow{4}{*}{UK BioBank \citeyear{Sudlow2015}} & \multirow{4}{*}{$18,926$} & \multirow{4}{*}{$12$} & \multirow{4}{*}{$5,000$} & \multirow{4}{*}{$500\,$Hz} \\
    {} & {} & {} & {} & {} & {} & {} \\
    {} & {} \\
    {} & {} & {} & {} & {} & {} & {} \\
    \midrule
    \multirow{8}{*}{\STAB{\rotatebox[origin=c]{90} {\footnotesize{Forecasting}}}} & \multirow{8}{*}{MSE}  & \multirow{4}{*}{Energy} & ETTh1 \citeyear{Zhou2021} & $1$ & $7$ & $17,420$ & (hourly) $278\,\mu$Hz \\
    {} & {} & {} & ETTh2 \citeyear{Zhou2021} & $1$ & $7$ & $17,420$ & (hourly) $278\,\mu$Hz \\
    {} & {} & {} & ETTm1 \citeyear{Zhou2021} & $1$ & $7$ & $69,680$ & (minutely) $1.1\,$mHz \\
    {} & {} & {} & ETTm2 \citeyear{Zhou2021} & $1$ & $7$ & $69,680$ & (minutely) $1.1\,$mHz \\
    {} & {} & Weather & Weather \citeyear{Wetter2024} & $1$ & $21$ & $52,696$ & (minutely) $2.8\,$mHz \\
    {} & {} & Electricity & Electricity \citeyear{UCI2024} & $321$ & $1$ & $26,304$ & (hourly) $278\,\mu$Hz \\
    {} & {} & Disease & Illness \citeyear{ILI2024} & $1$ & $7$ & $966$ & (weekly) $1.6\,\mu$Hz \\
    {} & {} & Logistics & Traffic \citeyear{PeMS2024} & $862$ & $1$ & $17,544$ & (hourly) $278\,\mu$Hz \\
    \bottomrule
\end{tabular}
\end{table}

\section{Baseline Details}
\label{sec:app_baseline_details}
Table \ref{tab:app_baseline_summary} provides an overview of all baselines evaluated in this study.

\begin{table}[!t]
\caption{
Summary of all baselines evaluated in our experiments, including details on architecture and pre-training strategy.
CL, MDM, and GPT refer to contrastive learning, masked data modelling, and autoregressive pre-training, respectively.
GP denotes general-purpose encoder.
}
\label{tab:app_baseline_summary}
\centering
\footnotesize
\setlength{\tabcolsep}{1.0em}
{\color{black}\begin{tabular}{llccc}
    \toprule
    \multirow{2}{*}{\textbf{Task}} & \multirow{2}{*}{\textbf{Model}} & \multirow{2}{*}{\textbf{Architecture}} & \multicolumn{2}{c}{\textbf{Pre-training}} \\
    \cmidrule(lr){4-5}
    {} & {} & {} & \textbf{Strategy} & \textbf{Dataset (Time points)} \\
    \midrule
    \multirow{15}{*}{\STAB{\rotatebox[origin=c]{90}{Classification}}}
        & \texttt{TS2Vec} \citeyear{Yue2022} & $1$D-CNN & CL & SleepEEG* ($74\,$M) \citeyear{Kemp2000} \\
        & \texttt{TimesNet} \citeyear{Wu2022} & $2$D-CNN & -- & -- \\
        & \texttt{TF-C} \citeyear{Zhang2022} & Transformer & CL & SleepEEG* ($74\,$M) \citeyear{Kemp2000} \\
        & \texttt{Ti-MAE} \citeyear{Li2023} & Transformer & MDM & SleepEEG* ($74\,$M) \citeyear{Kemp2000} \\
        & \texttt{SimMTM} \citeyear{Dong2023} & Transformer & MDM & SleepEEG* ($74\,$M) \citeyear{Kemp2000} \\
        & \texttt{PatchTST} \citeyear{Nie2023} & Transformer & -- & -- \\
        & \texttt{TSLANet} \citeyear{Eldele2024} & Transformer & -- & -- \\
        & \texttt{CroSSL} \citeyear{Deldari2024} & 1D-CNN & -- & -- \\
        & \texttt{InvConvNet} \citeyear{Germain2025} & 1D-CNN & -- & -- \\
        & \texttt{Handcrafted} \citeyear{Engemann2022} & Statistical & -- & -- \\
        & \texttt{ShallowNet} \citeyear{Schirrmeister2017} & 2D-CNN & -- & -- \\
        & \texttt{DeepNet} \citeyear{Schirrmeister2017} & 2D-CNN & -- & -- \\
        & \texttt{EEGNet} \citeyear{Lawhern2018} & 2D-CNN & -- & -- \\
        & \texttt{CBraMod} \citeyear{Wang2025CBra} & Transformer & MDM & TUEG ($127\,$B) \citeyear{Obeid2016} \\
        & \texttt{CSBrain} \citeyear{Zhou2025} & Transformer & MDM & TUEG ($127\,$B) \citeyear{Obeid2016} \\
    \midrule
    \multirow{8}{*}{\STAB{\rotatebox[origin=c]{90}{Regression}}}
        & \texttt{PatchTST} \citeyear{Nie2023} & Transformer & -- & -- \\
        & \texttt{ECG-Transformer} \citeyear{Dosovitskiy2021} & Transformer & -- & -- \\
        & \texttt{TSLANet} \citeyear{Eldele2024} & Transformer & -- & -- \\
        & \texttt{CroSSL} \citeyear{Deldari2024} & 1D-CNN & -- & -- \\
        & \texttt{InvConvNet} \citeyear{Germain2025} & 1D-CNN & -- & -- \\
        & \texttt{ECG-MAE} \citeyear{He2022} & Transformer & MDM & UK BioBank ($2.4\,$B) \citeyear{Sudlow2015} \\
        & \texttt{CM-AE} \citeyear{Radhakrishnan2023} & $1$D-CNN & MDM and CL & UK BioBank ($2.4\,$B) \citeyear{Sudlow2015} \\
        & \texttt{MMCL} \citeyear{Turgut2025} & Transformer & MDM and CL & UK BioBank ($2.4\,$B) \citeyear{Sudlow2015} \\
    \midrule
    \multirow{12}{*}{\STAB{\rotatebox[origin=c]{90}{Forecasting}}}
        & \texttt{N-BEATS} \citeyear{Oreshkin2019} & Non-Linear Model & -- & -- \\
        & \texttt{DLinear} \citeyear{Zeng2023} & Linear Model & -- & -- \\
        & \texttt{PatchTST} \citeyear{Nie2023} & Transformer & -- & -- \\
        & \texttt{GPT4TS} \citeyear{Zhou2023} & Transformer & GPT & $^\ddagger$ \\
        & \texttt{TTM} \citeyear{Ekambaram2024} & Transformer & MDM & Custom ($1\,$B) \\
        & \texttt{MOIRAI} \citeyear{Woo2024} & Transformer & MDM & Custom ($27\,$B) \\
        & \texttt{Chronos} \citeyear{Ansari2024} & Transformer & GPT & Custom ($84\,$B) \\
        & \texttt{Timer-XL} \citeyear{Liu2025} & Transformer & GPT & Custom ($56\,$B) \\
        & \texttt{Time-MoE} \citeyear{Shi2025} & Transformer & GPT & Custom ($309\,$B) \\
        & \texttt{ROSE} & Transformer & MDM & -- \\
        & \texttt{DTAF} & Transformer & MDM and KL & -- \\
    \midrule
    \multirow{2}{*}{\STAB{\rotatebox[origin=c]{90}{GP}}}
        & \texttt{UniTS} \citeyear{Gao2024} & Transformer & MDM & Custom ($36\,$B) \\
        & \texttt{MOMENT} \citeyear{Goswami2024} & Transformer & MDM & Custom ($1.2\,$B) \\
    \bottomrule
    \multicolumn{5}{l}{\footnotesize{* $371,055$ uni-variate, 2-seconds EEG recordings sampled at a frequency of $100$\,Hz.}} \\
    \multicolumn{5}{l}{\footnotesize{$^\ddagger$ Method uses a GPT$2$ \citeyear{Radford2018} encoder pre-trained on $10$ billion text tokens.}}
\end{tabular}}
\end{table}

\section{Frozen-Encoder Evaluation Details}
\label{sec:app_performance}
Following pre-training, we freeze all encoders (\texttt{OTIS}, \texttt{MOMENT}, and \texttt{UniTS}) and evaluate them under three protocols: prototypical $5$-shot classification, weighted $k$-nearest-neighbour classification, and linear probing.
For \texttt{OTIS} on any given downstream dataset, the variate embeddings are randomly initialised, and the encoder's output tokens are mean-pooled to yield a single dense representation per sample.

\paragraph{Prototypical $5$-shot classification.}
In each evaluation episode, $k\!=\!5$ training samples per class are randomly drawn to form the support set, and their representations are averaged to create a single prototype per class. 
Test samples are subsequently classified based on cosine similarity, assigning each sample to the nearest class prototype within the representation space. 
To reduce variance from the random selection of support samples, this prototypical $k$-shot procedure is repeated over $5$ independent episodes with freshly drawn support sets; balanced accuracy is reported as the mean $\pm$ standard deviation.
Random seeds are fixed to ensure that all three general-purpose encoders receive identical support sets in every episode.

\paragraph{Weighted $k$-nearest-neighbour classification.}
Following the \texttt{DINO} \citep{Oquab2024} protocols, the top-$k$ cosine neighbours are retrieved from the training representations for each test sample representation.
Each neighbour $i$ votes with weight $w_i = \exp\left( \cos(z, z_i) / \tau \right)$, where $z$ and $z_i$ denote representations of the test sample and the $i$-th neighbour, respectively. 
These weights are summed per class, and the final prediction is determined by the argmax over classes. 
We set $k\!=\!20$ and $\tau\!=\!0.07$ throughout our experiments, and report the balanced accuracy on the test split.

\paragraph{Linear probing.}
A multinomial $L_2$-regularised logistic regression is fit on $L_2$-normalised representations using the \texttt{lbfgs} solver from \texttt{scikit-learn} with $\mathtt{max\_iter}\!=\!1000$.
The regularisation strength $C$ is tuned over the grid $C\!\in\!\{10^{-2}, 10^{-1}, 1, 10, 10^{2}\}$ using a stratified $20\,\%$ holdout set. 
If fewer than $5$ samples per class or fewer than $50$ training samples are available, a $5$-fold stratified cross-validation is used instead. 
The regression model is then refit on the full training split with the optimal $C$, and balanced accuracy is reported on the test split.

\section{Deployment Evaluation Details}
\label{sec:app_deployment}
We measure all deployment metrics, including the memory, energy, latency, and throughput of each general-purpose encoder, on a single NVIDIA RTX A$6000$-$48$GB GPU. 
Inference is executed using \texttt{bfloat16} autocast (\texttt{torch.amp.autocast}), where weights are stored in \texttt{fp32} and matrix multiplications are performed in \texttt{bf16}. 
We use a fixed batch size of $B = 1024$, setting the sequence length $T$ and the number of variates $V$ to the native dimensions of each dataset. 
To suppress transient effects, we discard $10$ warm-up forward passes before averaging the measurements over $25$ timed iterations. 
The timed region encompasses the encoder's forward pass, including tokenisation; it explicitly excludes data loading, normalisation, and host-to-device transfers, as all tensors are moved to the GPU before timing commences. 
We rely on \texttt{time.perf\_counter} for wall-clock timing, \texttt{torch.cuda} for device synchronisation and peak-memory tracking, and NVIDIA NVML (\texttt{pynvml}) for board-level power measurements.

To ensure a fair comparison across encoders of different sizes, all four metrics are normalised to per-sample quantities, making them invariant to the chosen batch size $B$. 
We define the individual metrics as follows. 
\emph{Memory} [MB/Sample] is the peak allocation reported by \texttt{torch.cuda.max\_memory\_allocated} (following a call to \texttt{reset\_peak\_memory\_stats} after the warm-up phase) divided by the batch size $B$. 
\emph{Energy} [kWh/1M Samples] is computed by sampling the NVML board power before and after each iteration, calculating the mean iteration energy as $P_\text{mean} \cdot t_\text{elapsed}$, accumulating this across all $25$ iterations, and normalising per $1$ million samples.
\emph{Latency} [ms/Sample] is the mean elapsed time derived by invoking \texttt{torch.cuda.synchronize()} immediately before and after each forward pass, and dividing this by the batch size $B$.  
\emph{Throughput} [Samples/s] is the total number of samples processed over the $25$ iterations divided by the total elapsed time.

\section{Additional Results: Effect of Dual Masking on Time Series Feature Quality}
\label{sec:app_dual_masking}
We ablate the two hyperparameters that govern our dual masking strategy: the post-fix probability $p_\text{pf}$, which sets the fraction of pre-training steps that use post-fix masking rather than random masking, and the post-fix masking ratio $r_\text{pf}$, which sets the fraction of future time steps masked during post-fix steps.
Table~\ref{tab:app_masking_ablation} reports classification (balanced accuracy of prototypical 5-shot classification on the UCR archive) and forecasting (NCC and MSE on sine waves with a lookback window of $t_\text{past} = 1,400$ and predicted horizon of $t_\text{future} = 600$ time steps), averaged over three pre-training seeds, for perturbations around the default configuration.

Disabling post-fix masking entirely ($p_\text{pf}\!=\!0.0$, i.e. purely random masking) is the worst performing setting: the forecasting NCC drops from $0.957$ to $0.912$ and the MSE nearly doubles ($1.79 \to 3.57$), while classification accuracy falls by $1.5$\,pp ($59.24 \to 57.74\,\%$).
This mirrors the full-training ablation (Tab.~\ref{tab:ablation}), where reverting to purely random masking severely degrades forecasting performance. It confirms that exposing the encoder to a forecasting objective during pre-training is critical for learning temporal causality.
Raising the post-fix probability to $p_\text{pf}\!=\!0.5$ marginally improves classification ($+0.2$\,pp over the default) but at the cost of forecasting quality (NCC $0.948$, MSE $2.15$), as fewer steps are devoted to structural reconstruction.
Varying the post-fix masking ratio reveals a similar trade-off: a shorter forecasting horizon ($r_\text{pf}\!=\!0.25$) achieves classification performance comparable to the default but leads to less accurate forecasts, whereas a longer horizon ($r_\text{pf}\!=\!0.75$) hides too much context and harms both tasks ($57.56\,\%$, NCC $0.923$, MSE $3.10$).
The default configuration of $p_\text{pf}\!=\!0.25$ and $r_\text{pf}\!=\!0.5$ thus strikes the best overall balance, achieving near-best classification alongside the strongest forecasting performance, and is used throughout the study.

\begin{table}[!t]
\caption{
\textbf{Effect of dual masking on prototypical 5-shot UCR classification and zero-shot sine wave forecasting.}
For pre-training, post-fix probability $p_\text{pf}=0.25$ and post-fix masking ratio $r_\text{pf}=0.5$ are used by default. 
For zero-shot forecasting, sine waves span $2,000$ time steps, with a lookback window of $1,400$ time steps and a prediction horizon of $600$ time steps.
\textbf{Best} and \underline{second-best} scores are highlighted.
Mean and standard deviation are reported across three seeds set during pre-training.
}
\label{tab:app_masking_ablation}
\centering
\setlength{\tabcolsep}{0.55em}
\begin{tabular}{lccc}
    \toprule
    Masking configuration & Classification bACC $\uparrow$ & Forecasting NCC $\uparrow$ & Forecasting MSE $\downarrow$ \\
    \midrule
    \textcolor{gray}{\textit{Default}} \\ 
    \quad $p_\text{pf}{=}0.25$ and $r_\text{pf}{=}0.5$ & \underline{$59.24$} {\scriptsize$\pm 0.63$} & $\mathbf{0.957}$ {\scriptsize$\pm 0.01$} & $\mathbf{1.79}$ {\scriptsize$\pm 0.29$} \\
    \midrule
    \multicolumn{4}{l}{\textcolor{gray}{\textit{w/ different post-fix probability} $p_\text{pf}$}}\\
    \quad $p_\text{pf}{=}0.00$ (random masking only) & $57.74$ {\scriptsize$\pm 0.64$} & $0.912$ {\scriptsize$\pm 0.01$} & $3.57$ {\scriptsize$\pm 0.30$} \\
    \quad $p_\text{pf}{=}0.50$ & $\mathbf{59.45}$ {\scriptsize$\pm 0.58$} & \underline{$0.948$} {\scriptsize$\pm 0.01$} & \underline{$2.15$} {\scriptsize$\pm 0.22$} \\
    \midrule
    \multicolumn{4}{l}{\textcolor{gray}{\textit{w/ different post-fix masking ratio} $r_\text{pf}$}}\\
    \quad $r_\text{pf}{=}0.25$ & $59.20$ {\scriptsize$\pm 0.85$} & $0.946$ {\scriptsize$\pm 0.01$} & $2.21$ {\scriptsize$\pm 0.43$} \\
    \quad $r_\text{pf}{=}0.75$ & $57.56$ {\scriptsize$\pm 0.20$} & $0.923$ {\scriptsize$\pm 0.01$} & $3.10$ {\scriptsize$\pm 0.19$} \\
    \bottomrule
\end{tabular}
\end{table}

\section{Additional Results: Effect of NCC Loss on Time Series Feature Quality}
\label{sec:app_ncc}
We ablate the NCC loss weight $\lambda$ by evaluating $5$-shot prototypical classification performance across $6$ benchmark datasets.
Table~\ref{tab:ncc_ablation_kshot} reports the balanced accuracy for $\lambda \in \{0.0, 0.1, 0.2\}$. 
Setting $\lambda\!=\!0.1$ improves the overall average by $2.0$\,pp compared to $\lambda\!=\!0.0$ ($67.0\,\%$ vs. $65.0\,\%$). It achieves gains on $5$ out of the $6$ datasets, most notably on epilepsy ($+5.5$\,pp) and hand gesture ($+2.8$\,pp); only sleep stage drops marginally ($-0.8$\,pp).
Increasing the weight to $\lambda\!=\!0.2$ further benefits epilepsy ($+7.1$\,pp) and hand gesture ($+5.8$\,pp over $\lambda\!=\!0.0$), but degrades performance on sleep stage ($-2.6$\,pp) and muscular disease ($-1.1$\,pp).
These trends align with the full-training ablation (Tab.~\ref{tab:ablation}), where the removal of the NCC loss causes the largest performance drop, thereby motivating our choice of $\lambda\!=\!0.1$ throughout the study.

\begin{table}[!t]
\caption{
\textbf{Effect of NCC loss weight $\lambda$ on $5$-shot prototypical classification performance.}
\textbf{Best} and \underline{second-best} balanced ACC (in \% $\uparrow$) highlighted per dataset.
NCC $\lambda = 0.1$ provides best results across tasks.
}
\label{tab:ncc_ablation_kshot}
\centering
\footnotesize
\setlength{\tabcolsep}{0.85em}
\begin{tabular}{ccrrrccc}
\toprule
\multicolumn{5}{c}{\textbf{\normalsize Dataset}} & \multicolumn{3}{c}{\textbf{\normalsize Performance @ NCC $\lambda$}} \\
\cmidrule(lr){1-5} \cmidrule(lr){6-8}
\textbf{Domain} & \textbf{Name} & \textbf{Variates} & \textbf{Time points} & \textbf{Classes} & $\lambda\!=\!0.0$ & $\lambda\!=\!0.1$ & $\lambda\!=\!0.2$ \\
\midrule
Electromechanics & Machine fault   & $1$  & $5120$ & $3$ & $\underline{73.7}$ {\tiny$\pm 6.4$} & $\mathbf{75.4}$ {\tiny$\pm 5.8$} & $72.8$ {\tiny$\pm 3.4$} \\
Acceleration & Hand gesture    & $3$  & $206$  & $8$ & $54.7$ {\tiny$\pm 3.6$} & $\underline{57.5}$ {\tiny$\pm 1.9$} & $\mathbf{60.5}$ {\tiny$\pm 2.6$} \\
EMG          & Muscular disease   & $1$  & $1500$ & $3$ & $\underline{93.9}$ {\tiny$\pm 3.9$} & $\mathbf{95.9}$ {\tiny$\pm 0.7$} & $92.8$ {\tiny$\pm 4.4$} \\
\multirow{3}{*}{EEG} & Epilepsy & $1$  & $178$  & $2$ & $80.3$ {\tiny$\pm 5.1$} & $\underline{85.8}$ {\tiny$\pm 2.8$} & $\mathbf{87.4}$ {\tiny$\pm 2.6$} \\
{} & Sleep stage              & $1$  & $3000$ & $5$ & $\mathbf{47.9}$ {\tiny$\pm 5.6$} & $\underline{47.1}$ {\tiny$\pm 5.0$} & $45.3$ {\tiny$\pm 5.7$} \\
{} & Event type        & $19$ & $1000$ & $6$ & $39.4$ {\tiny$\pm 5.2$} & $\mathbf{40.5}$ {\tiny$\pm 6.5$} & $\underline{39.5}$ {\tiny$\pm 6.2$} \\
\midrule
\multicolumn{5}{c}{\textbf{Average}} & $65.0$ {\tiny$\pm 5.0$} & $\mathbf{67.0}$ {\tiny$\pm 3.8$} & $\underline{66.4}$ {\tiny$\pm 4.2$} \\
\bottomrule
\end{tabular}
\end{table}

\section{Additional Results: Prototypical Few-Shot Classification}
\label{sec:app_kshot}
Tables~\ref{tab:kshot_standard} and \ref{tab:kshot_cls2} report the prototypical $5$-shot classification performance per dataset for the standard benchmark and the UEA multivariate archive, respectively. Table~\ref{tab:kshot_overall} summarises the results across all classification datasets evaluated in our study.
Averaged across all datasets, \texttt{OTIS} matches the much larger \texttt{MOMENT} in accuracy ($59.3\,\%$ vs.\ $59.7\,\%$) and clearly outperforms \texttt{UniTS} which is of comparable size.
In terms of deployment efficiency, \texttt{OTIS} requires $10\!\times$ less memory ($3.5$ vs.\ $34.5$\,MB/sample), consumes $43\!\times$ less energy ($0.03$ vs.\ $1.32$\,kWh/1M\,samples), and achieves $37\!\times$ lower latency ($0.50$ vs.\ $18.5$\,ms/sample) than \texttt{MOMENT}, while delivering $8\!\times$ higher throughput ($3{,}574$ vs.\ $425$\,samples/s).
Although \texttt{UniTS} is the most efficient encoder ($1.6$\,MB, $0.21$\,ms, $10{,}844$\,samples/s), it trails both other encoders by roughly $8$\,pp in accuracy.
Overall, \texttt{OTIS} is the only encoder that achieves remarkable performance while maintaining deployment efficiency, indicating that its learned features are well suitable for high-quality, prototype-based diagnostics.

\begin{table}[!t]
\caption{
\textbf{Prototypical $5$-shot classification.}
\textbf{Best} and \underline{second best} balanced accuracy (bACC in \% $\uparrow$) are highlighted.
For each dataset, $V$, $T$, and $N$ denote the number of variates, time points, and classes, respectively.
\texttt{MOMENT} ($386$\,M) achieves the highest average $5$-shot accuracy but its memory and latency scale steeply with $T$ (up to $145$\,MB and $369$\,ms), making it impractical for deployment.
\texttt{UniTS} ($8.2$\,M) scales best with input size $V$ and $T$ but trails in performance.
In contrast, \texttt{OTIS} ($7.1$\,M) is wearable-feasible ($T\!\leq\!1500$, $V\!\leq\!3$: $<$$10$\,MB, $<$$4$\,ms), while ranking second in $5$-shot performance and dominating full fine-tuning.
}
\label{tab:kshot_standard}
\centering
\scriptsize
\setlength{\tabcolsep}{0.35em}
\begin{tabular}{llccccccc}
\toprule
\textbf{Dataset} & \textbf{Encoder} & \textbf{Full FT} & \textbf{$5$-Shot bACC} & \textbf{\makecell{Memory\\{\tiny [MB/Sample]} $\downarrow$}} & \textbf{\makecell{Energy\\{\tiny [kWh/1M\,Samples]} $\downarrow$}} & \textbf{\makecell{Latency\\{\tiny [ms/Sample]} $\downarrow$}} & \textbf{\makecell{Throughput\\{\tiny [Samples/s]} $\uparrow$}} \\
\midrule
        \multirow{3}{*}{\makecell[l]{\textbf{Machine fault}\\{\tiny ($V\!=\!1,T\!=\!5120,N\!=\!3$)}}} & \texttt{MOMENT-386M} & \underline{$92.8$} {\tiny$\pm 2.4$} & $\mathbf{69.0}$ {\tiny$\pm 7.2$} & $92.4$ & $25.59$ & $369.01$ & $2.7$ \\
                & \texttt{UniTS-8.2M} & $74.9$ {\tiny$\pm 5.2$} & $61.4$ {\tiny$\pm 2.7$} & $4.2$ & $0.18$ & $2.64$ & $378.3$ \\
                & \texttt{OTIS-7.1M} & $\mathbf{98.1}$ {\tiny$\pm 0.7$} & \underline{$62.6$} {\tiny$\pm 4.9$} & $9.9$ & $0.27$ & $3.84$ & $260.3$ \\
\midrule
        \multirow{3}{*}{\makecell[l]{\textbf{Hand gesture}\\{\tiny ($V\!=\!3,T\!=\!206,N\!=\!8$)}}} & \texttt{MOMENT-386M} & \underline{$61.8$} {\tiny$\pm 1.7$} & $\mathbf{53.3}$ {\tiny$\pm 4.0$} & $26.3$ & $1.64$ & $23.70$ & $42.2$ \\
                & \texttt{UniTS-8.2M} & $61.5$ {\tiny$\pm 0.9$} & $40.8$ {\tiny$\pm 4.3$} & $1.2$ & $0.02$ & $0.36$ & $2780.4$ \\
                & \texttt{OTIS-7.1M} & $\mathbf{78.0}$ {\tiny$\pm 2.0$} & \underline{$48.8$} {\tiny$\pm 3.0$} & $1.0$ & $0.04$ & $0.53$ & $1872.8$ \\
        \midrule
        \multirow{3}{*}{\makecell[l]{\textbf{Muscular disease}\\{\tiny ($V\!=\!1,T\!=\!1500,N\!=\!3$)}}} & \texttt{MOMENT-386M} & \underline{$97.6$} {\tiny$\pm 0.0$} & $\mathbf{94.6}$ {\tiny$\pm 1.0$} & $29.0$ & $1.00$ & $14.11$ & $70.9$ \\
                & \texttt{UniTS-8.2M} & $97.1$ {\tiny$\pm 1.1$} & $93.2$ {\tiny$\pm 2.4$} & $1.1$ & $0.01$ & $0.12$ & $8522.6$ \\
                & \texttt{OTIS-7.1M} & $\mathbf{97.6}$ {\tiny$\pm 0.0$} & \underline{$93.2$} {\tiny$\pm 2.8$} & $1.5$ & $0.02$ & $0.37$ & $2686.7$ \\
        \midrule
        \multirow{3}{*}{\makecell[l]{\textbf{Epileptic seizure}\\{\tiny ($V\!=\!1,T\!=\!178,N\!=\!2$)}}} & \texttt{MOMENT-386M} & \underline{$91.8$} {\tiny$\pm 0.7$} & $\mathbf{93.9}$ {\tiny$\pm 1.0$} & $23.3$ & $0.52$ & $7.43$ & $134.7$ \\
                & \texttt{UniTS-8.2M} & $87.8$ {\tiny$\pm 1.2$} & $86.4$ {\tiny$\pm 1.6$} & $0.6$ & $0.01$ & $0.10$ & $9741.2$ \\
                & \texttt{OTIS-7.1M} & $\mathbf{91.8}$ {\tiny$\pm 1.4$} & \underline{$92.1$} {\tiny$\pm 1.1$} & $0.7$ & $0.02$ & $0.26$ & $3780.3$ \\
        \midrule
        \multirow{3}{*}{\makecell[l]{\textbf{Sleep stage}\\{\tiny ($V\!=\!1,T\!=\!3000,N\!=\!5$)}}} & \texttt{MOMENT-386M} & $\mathbf{75.6}$ {\tiny$\pm 0.2$} & \underline{$33.9$} {\tiny$\pm 1.2$} & $29.8$ & $2.24$ & $33.16$ & $30.2$ \\
                & \texttt{UniTS-8.2M} & $71.0$ {\tiny$\pm 1.1$} & $31.4$ {\tiny$\pm 2.5$} & $2.3$ & $0.02$ & $0.24$ & $4155.0$ \\
                & \texttt{OTIS-7.1M} & \underline{$74.9$} {\tiny$\pm 0.6$} & $\mathbf{44.1}$ {\tiny$\pm 8.7$} & $4.0$ & $0.03$ & $0.42$ & $2375.5$ \\
        \midrule
        \multirow{3}{*}{\makecell[l]{\textbf{Event type}\\{\tiny ($V\!=\!19,T\!=\!1000,N\!=\!6$)}}} & \texttt{MOMENT-386M} & $51.7$ {\tiny$\pm 4.4$} & $\mathbf{53.3}$ {\tiny$\pm 3.0$} & $144.8$ & $10.31$ & $93.06$ & $10.7$ \\
                & \texttt{UniTS-8.2M} & \underline{$54.6$} {\tiny$\pm 2.1$} & $36.9$ {\tiny$\pm 5.3$} & $16.6$ & $0.06$ & $0.55$ & $1808.1$ \\
                & \texttt{OTIS-7.1M} & $\mathbf{61.4}$ {\tiny$\pm 1.3$} & \underline{$41.9$} {\tiny$\pm 4.7$} & $104.2$ & $0.19$ & $1.72$ & $582.4$ \\
        \midrule
        \multirow{3}{*}{\textbf{Average}} & \texttt{MOMENT-386M} & \underline{$78.6$} {\tiny$\pm 1.6$} & $\mathbf{66.3}$ {\tiny$\pm 2.9$} & $57.6$ & $6.88$ & $90.08$ & $48.6$ \\
                & \texttt{UniTS-8.2M} & $74.5$ {\tiny$\pm 1.9$} & $58.4$ {\tiny$\pm 3.1$} & $4.3$ & $0.05$ & $0.67$ & $4564.3$ \\
        \rowcolor{highlight} \cellcolor{white} & \texttt{OTIS-7.1M} & $\mathbf{83.6}$ {\tiny$\pm 1.0$} & \underline{$63.8$} {\tiny$\pm 4.2$} & $20.2$ & $0.10$ & $1.19$ & $1926.3$ \\
\bottomrule
\end{tabular}
\end{table}

\begin{table}[!t]
\caption{
\textbf{Prototypical $5$-shot classification on $23$ UEA archive datasets.}
\textbf{Best} and \underline{second best} balanced accuracy (bACC in \% $\uparrow$) highlighted.
\texttt{MOMENT} ($386$\,M) is powerful but unsuitable for deployment on wearables and edge devices (up to $531$\,MB and $353$\,ms).
\texttt{UniTS} ($8.2$\,M) scales best with $V$ and $T$ but trails in performance.
\texttt{OTIS} ($7.1$\,M) achieves the highest average $5$-shot performance while remaining wearable-feasible.
}
\label{tab:kshot_cls2}
\centering
\tiny
\setlength{\tabcolsep}{0.65em}
\begin{tabular}{llccccccc}
\toprule
\textbf{Dataset} & \textbf{Encoder} & \textbf{$5$-Shot bACC} & \textbf{Wins} & \textbf{\makecell{Memory\\ {\tiny [MB/Sample]} $\downarrow$}} & \textbf{\makecell{Energy\\ {\tiny [kWh/1M\,Samples]} $\downarrow$}} & \textbf{\makecell{Latency\\ {\tiny [ms/Sample]} $\downarrow$}} & \textbf{\makecell{Throughput\\ {\tiny [Samples/s]} $\uparrow$}} \\
\midrule
        \multirow{3}{*}{\makecell[l]{\textbf{EthanolConcentration}\\{\tiny ($V\!=\!3,T\!=\!1751,N\!=\!4$)}}} & \texttt{MOMENT-386M} & $24.3$ {\tiny$\pm 2.9$} & & $60.4$ & $19.7$ & $283.9$ & $3.5$ \\
                & \texttt{UniTS-8.2M} & \underline{$25.9$} {\tiny$\pm 2.3$} & & $4.7$ & $0.1$ & $1.6$ & $644.1$ \\
                & \texttt{OTIS-7.1M} & $\mathbf{26.4}$ {\tiny$\pm 3.8$} & & $10.1$ & $0.4$ & $5.4$ & $184.5$ \\
        \midrule
        \multirow{3}{*}{\makecell[l]{\textbf{Handwriting}\\{\tiny ($V\!=\!3,T\!=\!152,N\!=\!26$)}}} & \texttt{MOMENT-386M} & $\mathbf{18.5}$ {\tiny$\pm 0.2$} & & $25.3$ & $2.2$ & $31.6$ & $31.7$ \\
                & \texttt{UniTS-8.2M} & $10.7$ {\tiny$\pm 0.5$} & & $1.1$ & $0.0$ & $0.6$ & $1571.8$ \\
                & \texttt{OTIS-7.1M} & \underline{$11.1$} {\tiny$\pm 0.7$} & & $0.9$ & $0.0$ & $0.6$ & $1592.4$ \\
        \midrule
        \multirow{3}{*}{\makecell[l]{\textbf{JapaneseVowels}\\{\tiny ($V\!=\!12,T\!=\!26,N\!=\!9$)}}} & \texttt{MOMENT-386M} & \underline{$41.2$} {\tiny$\pm 3.4$} & & $24.0$ & $1.4$ & $19.6$ & $51.1$ \\
                & \texttt{UniTS-8.2M} & $38.8$ {\tiny$\pm 5.4$} & & $2.2$ & $0.1$ & $1.6$ & $633.8$ \\
                & \texttt{OTIS-7.1M} & $\mathbf{61.3}$ {\tiny$\pm 3.7$} & & $0.8$ & $0.0$ & $0.4$ & $2611.0$ \\
        \midrule
        \multirow{3}{*}{\makecell[l]{\textbf{SelfRegSCP1}\\{\tiny ($V\!=\!6,T\!=\!896,N\!=\!2$)}}} & \texttt{MOMENT-386M} & $62.7$ {\tiny$\pm 11.1$} & & $61.3$ & $19.5$ & $279.0$ & $3.6$ \\
                & \texttt{UniTS-8.2M} & \underline{$68.1$} {\tiny$\pm 7.5$} & & $5.1$ & $0.1$ & $1.8$ & $551.7$ \\
                & \texttt{OTIS-7.1M} & $\mathbf{68.5}$ {\tiny$\pm 4.9$} & & $10.6$ & $0.4$ & $5.5$ & $181.4$ \\
        \midrule
        \multirow{3}{*}{\makecell[l]{\textbf{SelfRegSCP2}\\{\tiny ($V\!=\!7,T\!=\!1152,N\!=\!2$)}}} & \texttt{MOMENT-386M} & $\mathbf{52.8}$ {\tiny$\pm 4.1$} & & $81.1$ & $24.5$ & $352.8$ & $2.8$ \\
                & \texttt{UniTS-8.2M} & \underline{$52.4$} {\tiny$\pm 2.6$} & & $7.2$ & $0.1$ & $1.9$ & $523.7$ \\
                & \texttt{OTIS-7.1M} & $50.0$ {\tiny$\pm 4.4$} & & $21.7$ & $0.6$ & $8.4$ & $119.6$ \\
        \midrule
        \multirow{3}{*}{\makecell[l]{\textbf{ArticularyWordRecognition}\\{\tiny ($V\!=\!9,T\!=\!144,N\!=\!25$)}}} & \texttt{MOMENT-386M} & \underline{$77.3$} {\tiny$\pm 1.7$} & & $31.5$ & $1.6$ & $22.6$ & $44.2$ \\
                & \texttt{UniTS-8.2M} & $60.0$ {\tiny$\pm 2.3$} & & $2.5$ & $0.0$ & $0.2$ & $4235.6$ \\
                & \texttt{OTIS-7.1M} & $\mathbf{79.0}$ {\tiny$\pm 0.9$} & & $1.8$ & $0.0$ & $0.3$ & $2992.2$ \\
        \midrule
        \multirow{3}{*}{\makecell[l]{\textbf{AtrialFibrillation}\\{\tiny ($V\!=\!2,T\!=\!640,N\!=\!3$)}}} & \texttt{MOMENT-386M} & \underline{$13.3$} {\tiny$\pm 0.0$} & & $103.0$ & $1.5$ & $20.6$ & $48.6$ \\
                & \texttt{UniTS-8.2M} & $6.7$ {\tiny$\pm 0.0$} & & $2.6$ & $0.0$ & $0.3$ & $3158.6$ \\
                & \texttt{OTIS-7.1M} & $\mathbf{33.3}$ {\tiny$\pm 0.0$} & & $3.6$ & $0.1$ & $1.0$ & $959.3$ \\
        \midrule
        \multirow{3}{*}{\makecell[l]{\textbf{BasicMotions}\\{\tiny ($V\!=\!6,T\!=\!100,N\!=\!4$)}}} & \texttt{MOMENT-386M} & \underline{$93.5$} {\tiny$\pm 1.2$} & & $39.3$ & $0.3$ & $4.8$ & $207.2$ \\
                & \texttt{UniTS-8.2M} & $75.0$ {\tiny$\pm 11.6$} & & $1.8$ & $0.0$ & $0.1$ & $7401.4$ \\
                & \texttt{OTIS-7.1M} & $\mathbf{100.0}$ {\tiny$\pm 0.0$} & & $1.4$ & $0.0$ & $0.4$ & $2582.6$ \\
        \midrule
        \multirow{3}{*}{\makecell[l]{\textbf{CharacterTrajectories}\\{\tiny ($V\!=\!3,T\!=\!180,N\!=\!20$)}}} & \texttt{MOMENT-386M} & $\mathbf{79.5}$ {\tiny$\pm 2.2$} & & $25.8$ & $0.3$ & $4.6$ & $217.4$ \\
                & \texttt{UniTS-8.2M} & $69.8$ {\tiny$\pm 2.2$} & & $1.2$ & $0.0$ & $0.1$ & $10540.1$ \\
                & \texttt{OTIS-7.1M} & \underline{$75.3$} {\tiny$\pm 1.2$} & & $0.9$ & $0.0$ & $0.2$ & $4179.7$ \\
        \midrule
        \multirow{3}{*}{\makecell[l]{\textbf{Cricket}\\{\tiny ($V\!=\!6,T\!=\!1197,N\!=\!12$)}}} & \texttt{MOMENT-386M} & $68.6$ {\tiny$\pm 1.9$} & & $74.4$ & $4.6$ & $67.9$ & $14.7$ \\
                & \texttt{UniTS-8.2M} & \underline{$79.7$} {\tiny$\pm 3.4$} & & $6.4$ & $0.0$ & $0.5$ & $2116.6$ \\
                & \texttt{OTIS-7.1M} & $\mathbf{86.7}$ {\tiny$\pm 1.9$} & & $17.0$ & $0.1$ & $1.0$ & $1003.5$ \\
        \midrule
        \multirow{3}{*}{\makecell[l]{\textbf{ERing}\\{\tiny ($V\!=\!4,T\!=\!65,N\!=\!6$)}}} & \texttt{MOMENT-386M} & $\mathbf{90.4}$ {\tiny$\pm 0.0$} & & $23.8$ & $0.2$ & $2.2$ & $461.9$ \\
                & \texttt{UniTS-8.2M} & $75.2$ {\tiny$\pm 0.0$} & & $1.1$ & $0.0$ & $0.1$ & $12271.6$ \\
                & \texttt{OTIS-7.1M} & \underline{$84.4$} {\tiny$\pm 0.0$} & & $0.7$ & $0.0$ & $0.2$ & $4108.4$ \\
        \midrule
        \multirow{3}{*}{\makecell[l]{\textbf{Epilepsy}\\{\tiny ($V\!=\!3,T\!=\!206,N\!=\!4$)}}} & \texttt{MOMENT-386M} & \underline{$79.5$} {\tiny$\pm 4.3$} & & $26.3$ & $0.7$ & $9.7$ & $103.3$ \\
                & \texttt{UniTS-8.2M} & $74.2$ {\tiny$\pm 5.2$} & & $1.2$ & $0.0$ & $0.1$ & $12281.8$ \\
                & \texttt{OTIS-7.1M} & $\mathbf{84.3}$ {\tiny$\pm 4.7$} & & $1.0$ & $0.0$ & $0.3$ & $3986.7$ \\
        \midrule
        \multirow{3}{*}{\makecell[l]{\textbf{FaceDetection}\\{\tiny ($V\!=\!144,T\!=\!62,N\!=\!2$)}}} & \texttt{MOMENT-386M} & \underline{$49.8$} {\tiny$\pm 1.0$} & & $147.2$ & $4.6$ & $65.9$ & $15.2$ \\
                & \texttt{UniTS-8.2M} & $49.5$ {\tiny$\pm 0.7$} & & $25.7$ & $0.2$ & $2.2$ & $457.6$ \\
                & \texttt{OTIS-7.1M} & $\mathbf{50.1}$ {\tiny$\pm 0.9$} & & $16.5$ & $0.1$ & $1.0$ & $1003.6$ \\
        \midrule
        \multirow{3}{*}{\makecell[l]{\textbf{FingerMovements}\\{\tiny ($V\!=\!28,T\!=\!50,N\!=\!2$)}}} & \texttt{MOMENT-386M} & $50.9$ {\tiny$\pm 0.4$} & & $31.8$ & $0.8$ & $11.8$ & $84.7$ \\
                & \texttt{UniTS-8.2M} & $\mathbf{52.3}$ {\tiny$\pm 2.7$} & & $5.3$ & $0.0$ & $0.4$ & $2428.5$ \\
                & \texttt{OTIS-7.1M} & \underline{$52.0$} {\tiny$\pm 5.0$} & & $1.9$ & $0.0$ & $0.2$ & $4083.3$ \\
        \midrule
        \multirow{3}{*}{\makecell[l]{\textbf{HandMovementDirection}\\{\tiny ($V\!=\!10,T\!=\!400,N\!=\!4$)}}} & \texttt{MOMENT-386M} & $24.4$ {\tiny$\pm 4.8$} & & $51.3$ & $2.6$ & $37.5$ & $26.7$ \\
                & \texttt{UniTS-8.2M} & $\mathbf{26.9}$ {\tiny$\pm 3.7$} & & $4.6$ & $0.0$ & $0.3$ & $3092.5$ \\
                & \texttt{OTIS-7.1M} & \underline{$26.3$} {\tiny$\pm 7.2$} & & $6.5$ & $0.0$ & $0.5$ & $1841.0$ \\
        \midrule
        \multirow{3}{*}{\makecell[l]{\textbf{Heartbeat}\\{\tiny ($V\!=\!61,T\!=\!405,N\!=\!2$)}}} & \texttt{MOMENT-386M} & $54.7$ {\tiny$\pm 7.3$} & & $530.8$ & $14.2$ & $203.5$ & $4.9$ \\
                & \texttt{UniTS-8.2M} & \underline{$57.1$} {\tiny$\pm 1.8$} & & $26.9$ & $0.3$ & $4.6$ & $217.2$ \\
                & \texttt{OTIS-7.1M} & $\mathbf{61.0}$ {\tiny$\pm 2.6$} & & $161.8$ & $0.3$ & $4.8$ & $206.6$ \\
        \midrule
        \multirow{3}{*}{\makecell[l]{\textbf{LSST}\\{\tiny ($V\!=\!6,T\!=\!36,N\!=\!14$)}}} & \texttt{MOMENT-386M} & $\mathbf{29.5}$ {\tiny$\pm 1.5$} & & $23.3$ & $0.1$ & $1.7$ & $577.2$ \\
                & \texttt{UniTS-8.2M} & $18.8$ {\tiny$\pm 1.2$} & & $1.4$ & $0.0$ & $0.1$ & $11051.6$ \\
                & \texttt{OTIS-7.1M} & \underline{$26.4$} {\tiny$\pm 3.2$} & & $0.7$ & $0.0$ & $0.3$ & $3973.6$ \\
        \midrule
        \multirow{3}{*}{\makecell[l]{\textbf{Libras}\\{\tiny ($V\!=\!2,T\!=\!45,N\!=\!15$)}}} & \texttt{MOMENT-386M} & \underline{$45.1$} {\tiny$\pm 2.4$} & & $22.5$ & $0.1$ & $0.8$ & $1276.7$ \\
                & \texttt{UniTS-8.2M} & $42.7$ {\tiny$\pm 3.5$} & & $0.7$ & $0.0$ & $0.1$ & $15857.9$ \\
                & \texttt{OTIS-7.1M} & $\mathbf{46.9}$ {\tiny$\pm 2.4$} & & $0.6$ & $0.0$ & $0.3$ & $3905.7$ \\
        \midrule
        \multirow{3}{*}{\makecell[l]{\textbf{NATOPS}\\{\tiny ($V\!=\!24,T\!=\!51,N\!=\!6$)}}} & \texttt{MOMENT-386M} & \underline{$50.2$} {\tiny$\pm 4.6$} & & $30.4$ & $0.7$ & $9.4$ & $106.0$ \\
                & \texttt{UniTS-8.2M} & $33.9$ {\tiny$\pm 3.2$} & & $4.6$ & $0.0$ & $0.4$ & $2823.1$ \\
                & \texttt{OTIS-7.1M} & $\mathbf{66.2}$ {\tiny$\pm 1.5$} & & $1.7$ & $0.0$ & $0.2$ & $4068.1$ \\
        \midrule
        \multirow{3}{*}{\makecell[l]{\textbf{Phoneme}\\{\tiny ($V\!=\!1,T\!=\!1024,N\!=\!39$)}}} & \texttt{MOMENT-386M} & \underline{$13.0$} {\tiny$\pm 0.4$} & & $29.4$ & $0.7$ & $9.6$ & $103.7$ \\
                & \texttt{UniTS-8.2M} & $12.4$ {\tiny$\pm 0.9$} & & $1.2$ & $0.0$ & $0.1$ & $11930.2$ \\
                & \texttt{OTIS-7.1M} & $\mathbf{16.3}$ {\tiny$\pm 0.4$} & & $1.6$ & $0.0$ & $0.3$ & $2924.1$ \\
        \midrule
        \multirow{3}{*}{\makecell[l]{\textbf{RacketSports}\\{\tiny ($V\!=\!6,T\!=\!30,N\!=\!4$)}}} & \texttt{MOMENT-386M} & \underline{$50.4$} {\tiny$\pm 5.6$} & & $23.0$ & $0.2$ & $2.7$ & $372.2$ \\
                & \texttt{UniTS-8.2M} & $40.2$ {\tiny$\pm 2.3$} & & $1.3$ & $0.0$ & $0.2$ & $5171.2$ \\
                & \texttt{OTIS-7.1M} & $\mathbf{57.3}$ {\tiny$\pm 3.2$} & & $0.7$ & $0.0$ & $0.2$ & $4151.4$ \\
        \midrule
        \multirow{3}{*}{\makecell[l]{\textbf{SpokenArabicDigits}\\{\tiny ($V\!=\!13,T\!=\!93,N\!=\!10$)}}} & \texttt{MOMENT-386M} & \underline{$53.6$} {\tiny$\pm 2.7$} & & $29.5$ & $1.3$ & $18.4$ & $54.4$ \\
                & \texttt{UniTS-8.2M} & $43.5$ {\tiny$\pm 3.9$} & & $3.0$ & $0.0$ & $0.2$ & $4544.1$ \\
                & \texttt{OTIS-7.1M} & $\mathbf{55.0}$ {\tiny$\pm 3.8$} & & $1.4$ & $0.0$ & $0.2$ & $4071.1$ \\
        \midrule
        \multirow{3}{*}{\makecell[l]{\textbf{UWaveGestureLibrary}\\{\tiny ($V\!=\!3,T\!=\!315,N\!=\!8$)}}} & \texttt{MOMENT-386M} & \underline{$60.6$} {\tiny$\pm 1.0$} & & $28.8$ & $1.2$ & $17.3$ & $57.9$ \\
                & \texttt{UniTS-8.2M} & $52.1$ {\tiny$\pm 2.0$} & & $1.5$ & $0.0$ & $0.2$ & $4942.6$ \\
                & \texttt{OTIS-7.1M} & $\mathbf{63.4}$ {\tiny$\pm 2.5$} & & $1.4$ & $0.0$ & $0.3$ & $3907.2$ \\
        \midrule
        \multirow{3}{*}{\textbf{Average}} & \texttt{MOMENT-386M} & \underline{$51.5$} {\tiny$\pm 2.8$} & \underline{$5$} & $66.3$ & $4.5$ & $64.3$ & $168.2$ \\
                & \texttt{UniTS-8.2M} & $46.3$ {\tiny$\pm 3.0$} & $2$ & $4.9$ & $0.1$ & $0.8$ & $5149.9$ \\
        \rowcolor{highlight} \cellcolor{white} & \texttt{OTIS-7.1M} & $\mathbf{55.7}$ {\tiny$\pm 2.6$} & $\mathbf{16}$ & $11.5$ & $0.1$ & $1.4$ & $2549.4$ \\
\bottomrule
\end{tabular}
\end{table}

\begin{table}[!t]
\caption{
\textbf{Average prototypical $5$-shot classification performance across benchmarks.}
\textbf{Best} and \underline{second best} balanced accuracy (bACC in \% $\uparrow$) are highlighted, with wins and deployment metrics reported alongside.
The \textbf{Overall} row aggregates metrics over all $149$ unique datasets (one UEA dataset overlaps with the standard benchmarks).
\texttt{OTIS} matches or even outperforms state-of-the-art encoders orders of magnitude larger, while saving $90\,\%$ memory, $90\,\%$ energy, and $97\,\%$ latency on average.
This marks a large step towards democratising access to high-quality time series features for any application and any use case.
}
\label{tab:kshot_overall}
\centering
\scriptsize
\setlength{\tabcolsep}{0.45em}
\begin{tabular}{llcccccc}
\toprule
\textbf{Benchmark} & \textbf{Encoder} & \textbf{$5$-Shot bACC} & \textbf{Wins} & \textbf{\makecell{Memory\\ {\scriptsize [MB/Sample]} $\downarrow$}} & \textbf{\makecell{Energy\\ {\scriptsize [kWh/1M\,Samples]} $\downarrow$}} & \textbf{\makecell{Latency\\ {\scriptsize [ms/Sample]} $\downarrow$}} & \textbf{\makecell{Throughput\\ {\scriptsize [Samples/s]} $\uparrow$}} \\
\midrule
\multirow{2}{*}{\makecell[l]{\textbf{Standard}\\{\tiny ($6$ datasets)}}}
        & \texttt{MOMENT-386M} & $\mathbf{66.3}$ {\tiny$\pm 2.9$} & $\mathbf{5}$ & $57.6$ & $6.88$ & $90.08$ & $48.6$ \\
        & \texttt{UniTS-8.2M} & $58.4$ {\tiny$\pm 3.1$} & $0$ & $4.3$ & $0.05$ & $0.67$ & $4564.3$ \\
        & \texttt{OTIS-7.1M} & \underline{$63.8$} {\tiny$\pm 4.2$} & \underline{$1$} & $20.2$ & $0.10$ & $1.19$ & $1926.3$ \\
\midrule
\multirow{2}{*}{\makecell[l]{\textbf{UEA}\\{\tiny ($23$ datasets)}}}
        & \texttt{MOMENT-386M} & \underline{$51.5$} {\tiny$\pm 2.8$} & \underline{$5$} & $66.3$ & $4.5$ & $64.3$ & $168.2$ \\
        & \texttt{UniTS-8.2M} & $46.3$ {\tiny$\pm 3.0$} & $2$ & $4.9$ & $0.1$ & $0.8$ & $5149.9$ \\
        & \texttt{OTIS-7.1M} & $\mathbf{55.7}$ {\tiny$\pm 2.6$} & $\mathbf{16}$ & $11.5$ & $0.1$ & $1.4$ & $2549.4$ \\
\midrule
\multirow{2}{*}{\makecell[l]{\textbf{UCR}\\{\tiny ($121$ datasets)}}}
        & \texttt{MOMENT-386M} & $\mathbf{60.9}$ {\tiny$\pm 3.3$} & $\mathbf{63}$ & $27.3$ & $0.437$ & $6.3$ & $492.2$ \\
        & \texttt{UniTS-8.2M} & $52.2$ {\tiny$\pm 3.5$} & $11$ & $0.9$ & $0.005$ & $0.1$ & $12238.2$ \\
        & \texttt{OTIS-7.1M} & \underline{$59.8$} {\tiny$\pm 3.4$} & \underline{$47$} & $1.1$ & $0.014$ & $0.3$ & $3850.6$ \\
\midrule
\multirow{2}{*}{\makecell[l]{\textbf{Overall}\\{\tiny ($149$ datasets)}}}
        & \texttt{MOMENT-386M} & $\mathbf{59.7}$ {\tiny$\pm 3.2$} & $\mathbf{73}$ & $34.5$ & $1.32$ & $18.50$ & $425.0$ \\
        & \texttt{UniTS-8.2M} & $51.5$ {\tiny$\pm 3.4$} & $13$ & $1.6$ & $0.02$ & $0.21$ & $10844.0$ \\
\rowcolor{highlight} \cellcolor{white} & \texttt{OTIS-7.1M} & \underline{$59.3$} {\tiny$\pm 3.3$} & \underline{$64$} & $3.5$ & $0.03$ & $0.50$ & $3574.0$ \\
\bottomrule
\end{tabular}
\end{table}

\clearpage
\section{Additional Results: Forecasting Visualisation}
\label{sec:app_forecast_vis}
\begin{figure}[!ht]
    \centering
    \includegraphics[width=1.0\linewidth]{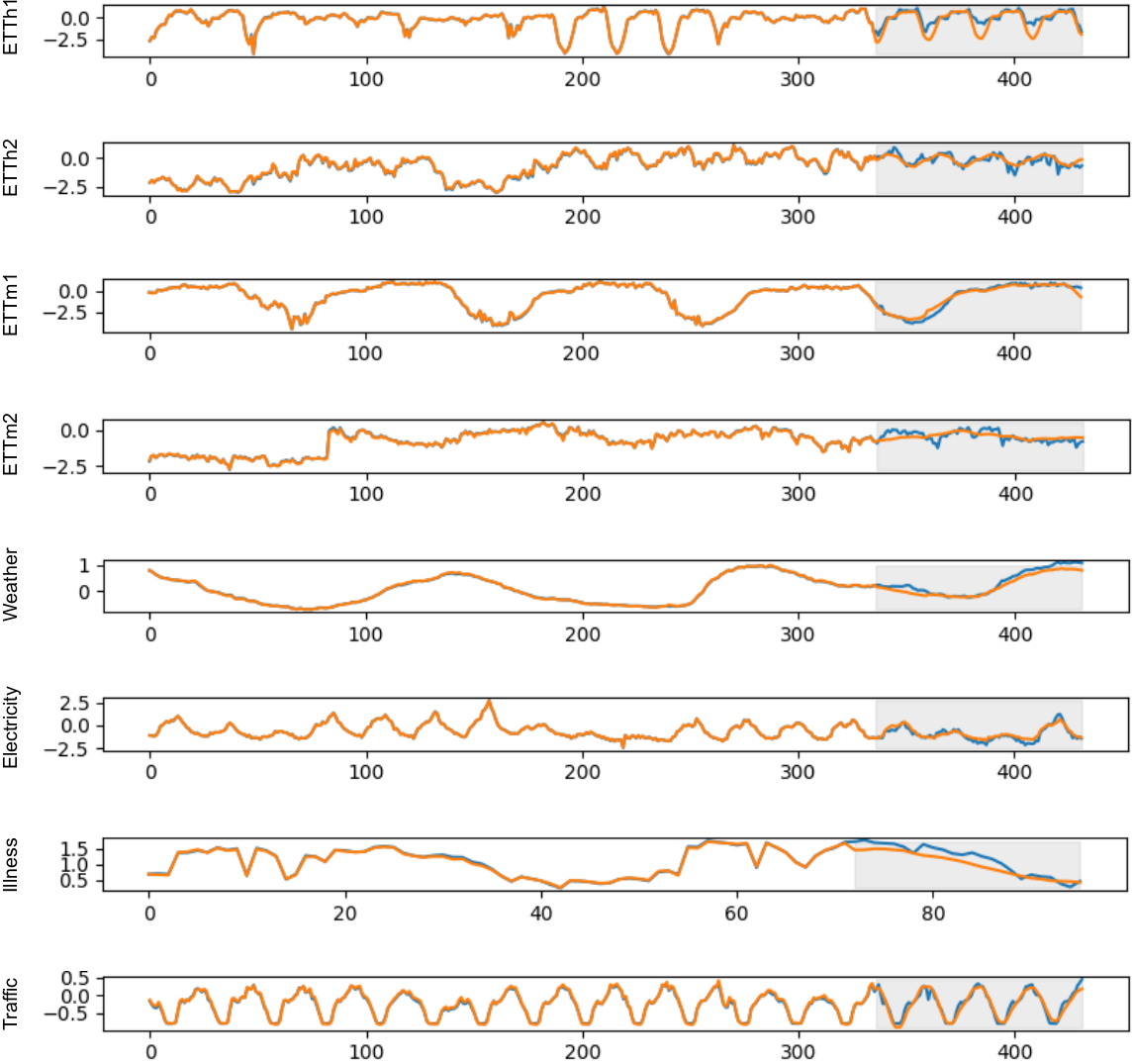}
    \caption{
    \textbf{Short-term forecasting using \texttt{OTIS} features.}
    Ground truth in \textcolor{RoyalBlue}{blue}, prediction in \textcolor{orange}{orange}.
    Areas in \textcolor{gray}{grey} are not visible to the encoder.
    \texttt{OTIS} features enable precise predictions of short horizons simply by reusing the lightweight $1.5\,$M decoder from pre-training.
    }
    \label{fig:app_forecast_vis}
\end{figure}

We visualise the predictions of \texttt{OTIS} on eight short-term forecasting datasets in Figure \ref{fig:app_forecast_vis}.

\end{document}